\documentclass[10pt,journal,compsoc]{IEEEtran}
\usepackage{graphicx}
\usepackage{textcomp}
\usepackage{amsmath} 
\usepackage{amsfonts}
\usepackage{amssymb}
\usepackage{wasysym}
\usepackage{bbm}
\usepackage{array}
\usepackage{xr}
\usepackage{verbatim}
\usepackage{float}
\usepackage{array,multirow}

\usepackage{booktabs}
\usepackage{graphicx,float}
\usepackage{subfig}
\usepackage[font={footnotesize}]{caption}
\usepackage{algorithm}
\usepackage{algpseudocode}

\begin{document}

\title{Learning the Regularization in DCE-MR Image Reconstruction for Functional Imaging of Kidneys}

\author{Aziz Ko\c{c}anao\u{g}ullar\i, {Cemre {Ariyurek}},
{Onur {Afacan}},
{Sila  {Kurugol}}
\thanks{Aziz Ko\c{c}anao\u{g}ullar{\i}, Cemre Ariyurek, Onur Afacan and Sila Kurugol are with Quantitative Intelligent Imaging Research Group (QUIN), Department of Radiology, Boston Children's Hospital and Harvard Medical School, Boston, MA, 02115, USA. (\{Aziz.Kocanaogullari, Cemre.Ariyurek, Sila.Kurugol, Onur.Afacan\}@childrens.harvard.edu)}
\thanks{This work was supported partially by the Society of Pediatric Radiology Multi-center Research Grant 2019, Crohn’s and Colitis Foundation of America’s (CCFA) Career Development Award and by the NIDDK and NIBIB of the National Institutes of Health under award numbers R01DK125561, R21DK123569, R21EB029627, and by the grant number 2019056 from the United States-Israel Binational Science Foundation (BSF), and a pilot grant from National Multiple Sclerosis Society under Award Number PP-1905-34002, and Society of Pediatric Radiology, Multicenter Research Award..}
\thanks{\textbf{\scriptsize{This work is accepted in IEEE Access (ID: Access-2021-40221, DOI: 10.1109/ACCESS.2021.3139854).}}}}

\IEEEtitleabstractindextext{
\begin{abstract}
Kidney DCE-MRI aims at both qualitative assessment of kidney anatomy and quantitative assessment of kidney function by estimating the tracer kinetic (TK) model parameters. Accurate estimation of TK model parameters requires an accurate measurement of the arterial input function (AIF) with high temporal resolution.   Accelerated imaging is used to achieve high temporal resolution, which yields under-sampling artifacts in the reconstructed images. Compressed sensing (CS) methods offer a variety of reconstruction options. Most commonly, sparsity of temporal differences is encouraged for regularization to reduce artifacts. Increasing regularization in CS methods removes the ambient artifacts but also over-smooths the signal temporally which reduces the parameter estimation accuracy. In this work, we propose a single image trained deep neural network to reduce MRI under-sampling artifacts without reducing the accuracy of functional imaging markers. Instead of regularizing with a penalty term in optimization, we promote regularization by generating images from a lower dimensional representation. In this manuscript we motivate and explain the lower dimensional input design. We compare our approach to CS reconstructions with multiple regularization weights. Proposed approach results in kidney biomarkers that are  highly correlated with the ground truth markers estimated using the CS reconstruction which was optimized for functional analysis. At the same time, the proposed approach reduces the artifacts in the reconstructed images.   
\end{abstract}

}

\maketitle
\section{Introduction}
\label{sec:intro}

Quantitative dynamic contrast-enhanced MRI (DCE-MRI) calculates perfusion and permeability parameters of tissue by evaluating the T1 shortening induced by a gadolinium-based contrast agent (CA) passing through the tissue of interest. A dynamic series of T1 weighted volumes are acquired before and after bolus injection of CA. A detailed anatomical evaluation is then performed by assessing the contrast uptake visually and a functional evaluation can be performed by fitting a tracer kinetic model and estimating its parameters. In functional imaging of the kidneys, DCE-MRI can be used for both quantitative and qualitative assessment of renal function, i.e., glomerular filtration rate (GFR) and differential renal function (DRF). A rapid acquisition with high temporal resolution ($\approx$3~seconds/volume) is required to capture the passage of CA in vascular system and through the kidneys for accurate computation of TK model parameters \cite{kurugol2020prospective, kurugol2020feed}. 

High temporal resolution can be achieved using accelerated imaging by taking advantage of parallel imaging and by undersampling the k-space using Cartesian sampling \cite{block2007undersampled}. Alternatively, accelerated non-Cartesian sampling such as stack-of-stars radial imaging can be used, which has an additional benefit of being inherently robust to breathing motion \cite{block2014towards}. When using accelerated imaging, image quality and quantitative parameter estimation accuracy are both degraded by undersampling artifacts, i.e streaking artifacts and noise~\cite{ravishankar2010mr}. In order to reduce the effect of these artifacts, compressed sensing (CS) methods have been used in the literature. For CS reconstruction, an incoherent sampling of the k-space is required, which can be achieved by golden angle stack-of-stars acquisition of dynamic series of volumes. In addition, DCE-MR data over time have spatiotemporal redundancy and this apriori information can be utilized during image reconstruction. A regularization term is usually linearly appended to a data consistency term with a tuning hyperparameter and then minimized in the optimization \cite{zhang2015accelerating, hu2020multi}. Literature offers several regularization terms, e.g., promoting low rank background \cite{otazo2015low}, promoting sparsity in time or space \cite{ong2016beyond}, promoting sparsity in extra dimensions by splitting data into respiratory phases \cite{feng2016xd}. Most commonly used regularization function for DCE-MRI is the one that promotes sparsity in temporal difference of the signal in time \cite{feng2014golden}. One drawback of this regularization scheme is that, increasing  the regularization hyperparameter reduces the unwanted streaking artifacts in the image improving image quality but this comes at the cost of reducing the tracer kinetic model parameter estimation accuracy due to over-smoothing of the temporal signal, especially that of the arterial input function. Therefore, for functional evaluation, a smaller regularization hyperparameter is selected when TK model fitting is required. However,  when a smaller regularization weight is used, the resultant images have more streaking artifacts, reducing the image quality. It is therefore desirable to have a DCE-MR image reconstruction approach that can achieve both good image quality and accurate TK parameter estimation at the same time.

Neural network architectures have gained popularity in the literature \cite{schlemper2017deep,liu2021magnetic} due to their ability to achieve regularization by learning the transform parameters without the need for hand-crafting the regularization function. Neural networks offer flexibility to determine the spatio-temporal redundancy in the dynamic MR images \cite{ke2020deep, han2018deep}. Deep learning techniques have shown great promise in reconstructing high quality images from accelerated MRI data. However, these techniques suffer from scalability. In clinical practice, collecting a representative training dataset is a difficult objective due to various sources of variations including different ages of patients, presence of motion, anatomical and pathological variations, variations in field of views and coil selection and placement. Additionally, deep learning based image reconstruction pipelines are usually trained in an end-to-end supervised fashion, in which under-sampled data are reconstructed with the network and compared to a ground-truth reference reconstruction. The \textit{reference} reconstruction usually refers to a reconstruction with k-space data that is fully sampled (or sampled denser than expected input) with a conventional technique or with a compressed sensing algorithm. The notion of reference is therefore tied to an already existing method, which may not be an ideal solution itself, and acquisition of such reference data suffers from similar difficulties or limitations such as limitations of temporal resolution. To avoid such drawbacks, a possible solution is to optimize the neural network for single image reconstruction without using a reference ground truth image. Ulyanov et al. in their work "Deep image-prior (DIP)" \cite{ulyanov2018deep} introduced the idea of reconstructing a single artifact free image by imposing a deep neural network regularizer. DIP is trained promoting fidelity between the deteriorated observation and network generated regularized image. Through iterations, with a sufficiently deep architecture, the network eventually overfits to the noisy observation itself. However, the intermediate steps yield parameter sets that result in repaired (de-noised, inpainted...), regularized reconstructions. 
Also, this approach is conducted on a single image, without any need for training data, by eliminating the learning process from a training set of input, output image pairs that may not generalize well to unseen data. 


In this work we propose an image  reconstruction algorithm for 3D+t (4D) DCE-MRI data that is based on a single subject trained regularizer neural network given the multichannel undersampled kspace data acquired at each time point. The proposed method does not require separate supervised training using a fully sampled training data set, which is not available. Instead of employing a regularization term in an optimization scheme, we learn the regularized reconstruction with a network architecture.  Coherent low-rank/sparse nature of MR images are assumed to be disturbed due to presence of undersampling artifacts and noise and the original image can be recovered from a corrupted image using a generator network. This network takes a randomly initialized lower dimensional input and generates an improved quality image. Our main contribution is expanding this regularization network developed for a 2D volume to our 3D+time (4D) DCE-MRI data, where we reconstruct each 2D slice independently and also integrate the temporal correlation of images when training the reconstruction algorithm. Instead of using a temporal series of lower dimensional random inputs, we design a series of random inputs which are correlated in time, mimicking the fact that DCE-MRI sequence of images are temporally correlated. For instance, in our application of kidney DCE-MRI, there is a no-contrast phase, an arterial phase (where aorta and renal arteries light up), a cortical phase, a medullary phase, and a pelvis phase (where respective kidney compartments light up). We design a lower dimensional input sequence utilizing {noisy inverse NUFFT reconstructions} to reflect the temporal correlation of a dynamic series of contrast enhanced images within each phase. Given this lower dimensional input sequence, and the corrupted sequence of DCE-MRI images, our generator network parameters are optimized to reconstruct a sequence of DCE-MR images with improved image quality spatially, while at the same time preserving the temporal dynamics of the contrast.

We extensively evaluate the quality of regularized reconstruction in terms of both the accuracy of estimated quantitative parameters and the quality of resultant images.
The contributions of this paper can be listed as follows:
(i) Introducing a deep neural network based regularizer for 4D DCE-MR image reconstruction that utilizes the spatio-temporal redundancy in the data. Designing a lower dimensional temporal sequence of images, which is provided as input to the regularizer network. 
(ii) The ability to train without requiring a fully sampled ground truth data and a training set of ground truth, artifact free DCE-MRI images.
(iii) Extensive evaluation of the proposed DCE-MR image reconstruction algorithm in patient data in terms of 1) spatial image quality, 2) temporal dynamics of the arterial input function and kidney signal quality and 3) quantitative imaging markers of kidney function estimation accuracy in comparison to a state of the art CS technique.

\section{Method}
\label{sec:theory}
In this section, 
we first introduce the problem formulation and notation. In Section~\ref{subsec:lower_dim} we explain the lower dimensional pilot input design for DCE-MRI. We then continue in Section~\ref{subsec:reg_network} with the architecture and training details of the learned regularization. This section is concluded with experimental setup description in Section~\ref{subsec:exp_setup}.

\label{subsec:prelim}
\subsection{DCE-MRI with stack-of-stars sampling} In DCE-MRI acquisition with stack-of-stars golden angle radial sampling of k-space, each radial line (spoke) is acquired for all slices followed by the next radial line acquired by rotating the previous line by the golden angle. K-space is dynamically populated with the respective observed radial samples. A dynamic series of complex valued volumes are then reconstructed at each time point $t$. After applying the Fourier transform in slice direction, each image slice $x_t\in\mathbb{C}^{n\times n}$ can be independently reconstructed. An image is related with frequency observations through non-uniform Fourier transform (NUFFT) denoted by $F$. K-space observations are acquired from multiple coils with the coil profile denoted by $C$. We denote frequency domain measurements for a time instance image $x_t$ as $k_t\in\mathbb{C}^{n\times c \times sp}$ where $n,c,sp$ denote frequency resolution, number of coils and number of spokes, respectively. 

\subsection{Compressed Sensing for DCE-MRI reconstruction}
\label{subsubsec:cs_methods}
Let $X=\lbrace x_t \lvert t\in \lbrace 1,2,\cdots,T\rbrace\rbrace$ denote DCE-MR image sequence.  A back projection solution is referred to as inverse-NUFFT (i-NUFFT) solution $\hat{x_t} = (FC)^{-1}k_t$. However, when using under-sampled k-space data, this approach is prune to streaking reconstruction artifacts. To avoid these artifacts different regularization terms have been proposed in CS literature that enforce sparsity and low rank structure in both spatial and temporal domains for dynamic image reconstruction. Regularized optimization scheme is noted in the following equation.
%
\begin{equation}
    \label{eq:reg_least_squares}
    \hat{X} = \arg\min_{X } \sum_{t\in\lbrace 1,\cdots, T\rbrace} \| FCx_t - k_t \|_2^2 + \sum\limits_i \lambda_i R_i(X)
\end{equation}

In dynamic reconstruction, with the assumption of stationary nature of consequent images, temporal piece-wise constancy of voxels are promoted with a temporal total variation regularization $R(X) = \| \sum_{t=\lbrace 1,\dots, T-1 \rbrace} | x_{t+1} - x_t |\|_{1,1}$ where $\| . \|_{1,1}$ is the entry wise $\ell_1$-norm. Sparsity of temporal changes is adjusted with the hyperparameter $\lambda$.

\subsection{Deep Image Prior (DIP) for regularization} Instead of designing a regularization function to reconstruct the image sequence $X$, we learn a regularized deep image generator network $f_\theta$ given a lower-dimensional input. Our approach is motivated by the Deep Image Prior technique (DIP) \cite{ulyanov2018deep}. DIP is a convolutional neural network (CNN) used to recover an  improved quality, regularized image from a random input image of the same size without using any prior training data other than the image itself. Moreover, such technique can use a lower dimensional input which is passed through the CNN to generate a regularized image and hence enforce a low-rank/sparse structure spatially. For dynamically acquired images, a simple correlated lower dimensional input sequence can enforce sparsity through time \cite{yoo2021time}.

For MR image reconstruction, we start from a lower dimensional representation $z_t\in\mathbb{R}^m$ of each image $x_t$. The low-rank/sparse nature of the reconstructed image is ensured with the dimensionality of the input $m\ll n\times n$. We design a neural network architecture $f_\theta$ to generate reconstructions $f_{\theta}(z_t) = x_t \in \mathbb{C}^{n\times n}$. Assuming a Gaussian observation model, the kspace measurements $k_t$s are noisy $FC(x_t)$ observations. Thus, it is possible to estimate the optimal parameter set $\theta$ for the architecture $f$ minimizing the squared $\ell_2$ loss between the transform of the generated image $FC(f_\theta(z_t))$ and the observation $k_t$ for each image of the dynamic sequence.

\begin{equation}
    \label{eq:loss_DIP}
    \min_{\theta} \frac{1}{T}\sum_{t\in\lbrace1,\cdots,T \rbrace}\| FC f_\theta(z_t )- k_t \|_2^2
\end{equation}  

Learned regularization parameters $(\theta)$ rely on the lower dimensional input $(z_t)$ and the optimization scheme. We describe these aspects in the following sections.

\subsection{Lower Dimensional Input Design}
\label{subsec:lower_dim}

%
\begin{figure}[t]
		\centering
		\includegraphics[width=.6\columnwidth]{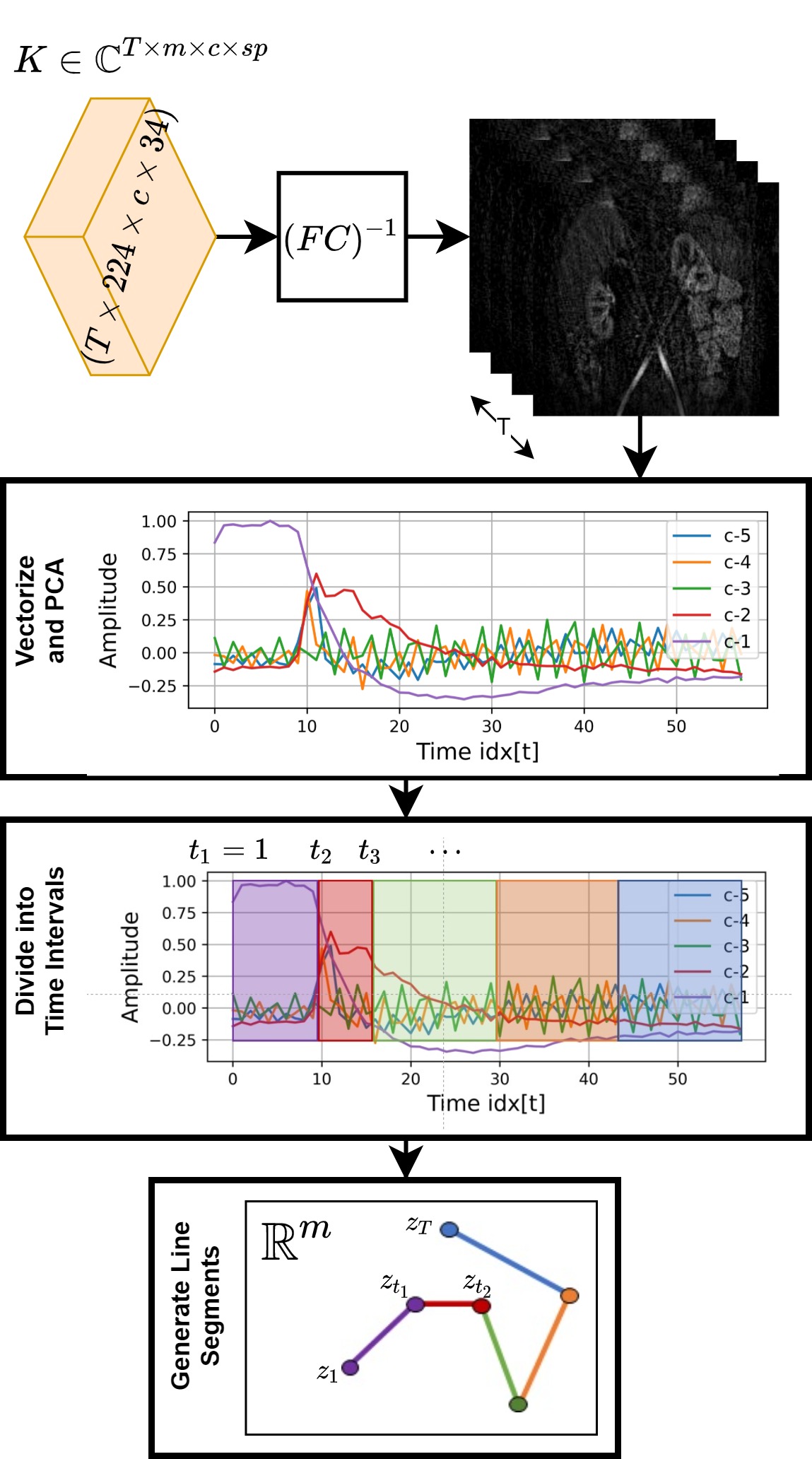}
      
      \caption{
      \label{fig:z_deisgn}
      Illustration of the proposed algorithm to generate a low dimensional representations ($z_t$'s) of the image sequence that is given as the input to the network and used to reconstruct artifact free image sequence ($x_t$'s). Illustration of how we determine the  segments of the dynamic signal:  We first apply the i-NUFFT to the k-space data from a slice to reconstruct an initial sequence of images from each coil. Absolute value of these images are vectorized to generate feature vectors. Singular value decomposition is then applied on these vectors to find most significant 5 components and respective feature projections. From the feature map, time intervals are determined with thresholding as shown in different colors. These intervals are used to generate connected line segment structure $z_t$'s with Algorithm~\ref{alg:z_design}. 
      }
\end{figure}

\begin{algorithm}[t]
\scriptsize
    \caption{Lower dimensional input design}
    \label{alg:z_design}
    \begin{algorithmic}[1] 
        \State \textbf{initialize} $\mu_z\in \mathbb{R}^m$, $\mu_\alpha\in \mathbb{R}^m$
        \Statex \hspace{0.2cm}$\ell\in\mathbb{R}^+$ \Comment{line segment length} 
        \Statex \hspace{0.2cm}$\lbrace t_1=1,t_2,t_3,t_4,t_5,t_6=T \rbrace$ \Comment{ordered indices }
        \Statex \hspace{0.3cm} s.t. $1\leq t_{i-1}<t_i\leq T , t_i \in \mathbb{Z}^+ \ \forall i$ 
        \State $\text{tmp}_z \gets \mathcal{N}(\mu_z, I)$ \Comment{randomly determine start}
        \State $z_1 \gets \text{tmp}_z$ \Comment{Set $z_1$}
        \For{$i_1\in \lbrace 1,2,3,4,5 \rbrace$} \Comment{all intervals}
            \State $\alpha \gets \mathcal{N}(\mu_\alpha, I)$ s.t. $\alpha \in \mathbb{R}^{m\times 1}$ \Comment{set random angle $\in\mathbb{R}^m$}
            \State $\text{idx}\gets 1$
            \For{$i_2 \in \lbrace t_{i_1}+1 ,t_{i_1}+2,\cdots, {t_{i_1+1}} \rbrace$} \Comment{all points in interval}
                \State $z_{i_2}\gets \text{tmp}_z + \alpha \times \left(\ell \frac{idx}{ t_{i_1+1}-t_{i_1}}\right) $ \Comment{equidistant points}
                \State $\text{idx} \gets\text{idx}+1$
            \EndFor
        \EndFor
        \State \textbf{return:} $z_t$ $\forall t\in\lbrace 1,2,\cdots,T \rbrace$
    \end{algorithmic}
\end{algorithm}

%
\begin{figure*}[t]
		\centering
		\includegraphics[width=1.6\columnwidth]{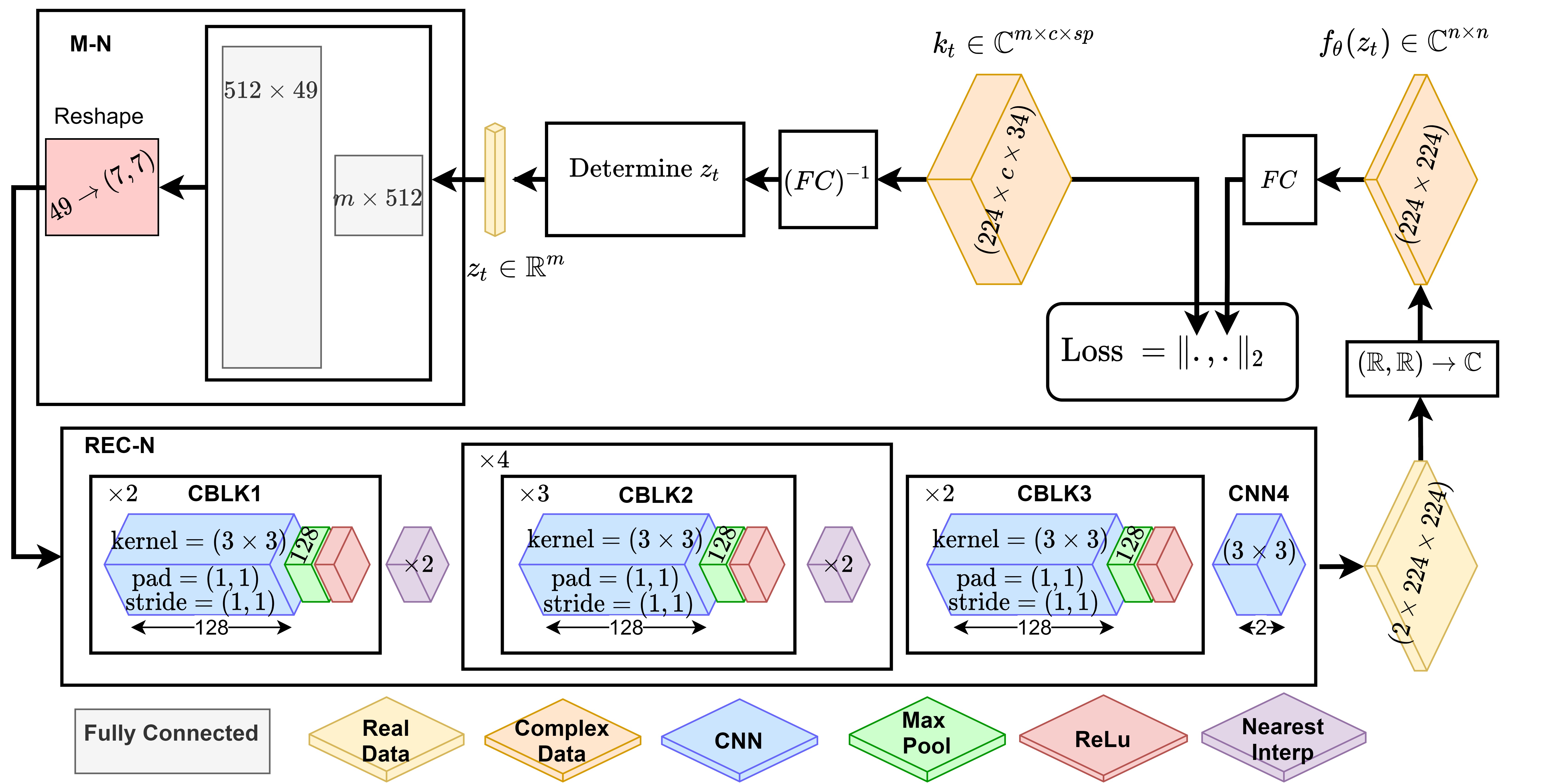}
\caption{ The architecture of the regularization network $f_\theta$ used to reconstruct each image $f_\theta(z_t)$ given the lower dimensional representation vector $z_t$ is shown. Observe that the network employs only real valued weights and outputs 2 channels both with size $(224\times 224)$ to represent the real and imaginary parts of the complex image. Let $a,b$ denote these two separate channels of the output. These channels are fused to form the complex image $f_\theta(z_t)= a + ib$. Generated images $f_\theta(z_t)$ are further passed through the forward model $FC$ to compute the $\ell_2$ loss with $k_t$.
\label{fig:gen_net}
}
\end{figure*}
\begin{table}[t]
\caption{Parameters for neural network architecture.\label{tab:network_param} \vspace{-0.3cm}}
{\def\arraystretch{0.3}\tabcolsep=4pt
    \begin{center}
        \footnotesize
	\begin{tabular}{c c | c c c c c}
            &Layer & Filter Size & Stride & 0-Pad & Output \\
            \midrule
            \parbox[t]{2mm}{\multirow{12}{*}{\rotatebox[origin=c]{90}{REC-N \hspace{1cm}}}} 
            &CNN4 &$(3\times 3\times 2)$ &$(1,1)$ &$(1,1)$ &$(224\times224\times2)$\\
            &CBLK3$\times 2$ &$(3\times 3\times 128)$ &$(1,1)$ &$(1,1)$ &$(224\times224\times128)$\\
            &I2 &$-$ &$(2,2)$ &$-$ &$(224\times224\times128)$\\
            &CBLK2$\times 3$ &$(3\times 3\times 128)$ &$(1,1)$ &$(1,1)$ &$(112\times112\times128)$\\
            &I2 &$-$ &$(2,2)$ &$-$ &$(112\times112\times128)$\\
            &CBLK2$\times 3$ &$(3\times 3\times 128)$ &$(1,1)$ &$(1,1)$ &$(56\times56\times128)$\\
            &I2 &$-$ &$(2,2)$ &$-$ &$(56\times56\times128)$\\
            &CBLK2$\times 3$ &$(3\times 3\times 128)$ &$(1,1)$ &$(1,1)$ &$(28\times28\times128)$\\
            &I2 &$-$ &$(2,2)$ &$-$ &$(28\times28\times128)$\\
            &CBLK2$\times 3$ &$(3\times 3\times 128)$ &$(1,1)$ &$(1,1)$ &$(14\times14\times128)$\\
            &I1 &$-$ &$(2,2)$ &$-$ &$(14\times14\times128)$\\
            &CBLK1$\times 2$ &$(3\times 3\times 128)$ &$(1,1)$ &$(1,1)$ &$(7\times7\times128)$\\
            \midrule
            \parbox[t]{2mm}{\multirow{3}{*}{\rotatebox[origin=c]{90}{M-N}}} 
            &Reshape &$-$ &$-$ &$-$ & $(7\times 7 \times 1)$\\
            &FC2 &$512\times49$ &$-$ &$-$ & $(49\times 1$)\\
            &FC1 &$m\times512$ &$-$ &$-$ & $(512 \times 1)$
    \end{tabular}
    \end{center}
    }
\end{table}

This section summarizes how we design a lower dimensional input that is used to generate a regularized reconstruction of the output image sequence. 
In DCE-MRI, dynamic sequence of images are temporally correlated. Therefore, we design a lower dimensional input sequence $Z = \lbrace z_t| \forall t\in\lbrace1,\cdots,T \rbrace\rbrace$ to represent the temporal dynamics of the contrast enhanced image sequence. 
The goal is to generate a sequence of images with improved image quality, while at the same time preserving the temporal dynamics of the signal {from lower dimensional input}.

It is possible to separate DCE-MR images temporally into multiple phases such as no contrast phase, arterial phase, and phases where different tissues enhance. 
Due to different phases, we propose to represent each phase with a line and hence we propose to use a piece-wise linear $Z$ sequence. By nature, each phase duration in DCE-MRI is scan specific, thus each line segment in $Z$ will have different number of samples. In this section, we propose an algorithm to assign each time point to its corresponding phase and generate the piece-wise linear $Z$ accordingly.
We utilize computationally inexpensive vectorized i-NUFFT reconstructions $|\text{vec}((FC)^{-1}(k_t))|$ for each time point $t$. Furthermore to reduce complexity, we use feature maps that correspond to top-n singular values of the vectorized i-NUFFT reconstructions. Singular value decomposition (SVD) based feature maps are shown to inhibit indicators of activity in medical imaging \cite{doneva2008automatic,zhang2013coil}. Therefore by simply thresholding the feature maps, we cluster consecutive time points to temporal phases. i-NUFFT driven thresholding with SVD for phase assignment is illustrated in Fig.\ref{fig:z_deisgn}. Once clusters are obtained, each time-index cluster is assigned to a line-segment where the earliest and latest time-indices are assigned as the two end points. The initial point of the first line segment is randomly sampled from an $m$-dimensional multivariate Gaussian distribution (where $m$ is the dimensionality of the lower dimensional representation). Our algorithm then samples an orientation from a multivariate Gaussian distribution for the equi-length line segments.  Once orientation is determined each point in the respective cluster is uniformly placed on the line segment. The algorithm then proceeds by generating the following line segments with an independently sampled random orientation for which the starting point of the line is the predecessor's end point. A sample generated piecewise linear curve in $\mathbb{R}^2$ is illustrated in Figure~\ref{fig:z_deisgn}. {We summarize generating lower dimensional input from the clustered time points in Algorithm~\ref{alg:z_design}.}

\subsection{Regularization Network}
\label{subsec:reg_network}

Our proposed approach utilizes a neural network $f_\theta$ to generate reconstructions $f_{\theta}(z_t) = x_t \in \mathbb{C}^{n\times n}$ ($z_t\in \mathbb{R}^m, \ m<< n\times n$). The architecture consists a modification network (M-N) and the convolutional reconstruction network (REC-N). We mentioned how to reflect the application nature in the lower dimensional representation in Section~\ref{subsec:lower_dim}. As expected, CA's physiological behavior employs a more complex structure than $Z$, therefore with a fully connected M-N we learn the required adjustments on the lower dimensional input. Modified input is then used to reconstruct the image through fully convolutional REC-N. The architecture is illustrated in Figure~\ref{fig:gen_net} and the details are presented in Table~\ref{tab:network_param}. Parameters $\theta$ are updated with batch gradient descent,

\begin{equation}
    \theta^{s+1} \gets \theta^{s} + \gamma\left. \frac{\delta \left(\frac{1}{|B|}\sum_{t\in B }\| FC f_\theta(z_t )- k_t \|_2^2\right)}{\delta\theta} \right|_{\theta = \theta^s}
\end{equation}

Here $B \subset \lbrace1,\cdots,T \rbrace$ denotes the batch with cardinality $|B|$, $\gamma$ denotes the learning rate and $s$ denotes the iteration number. The network will overfit to the undersampled k-space data by design. Thus,  different parameters $\theta^s$s through the iterations propose a family of reconstruction networks. The intermediate parameter values present different quality reconstructions, ultimately allowing the flexibility to determine a regularization between different parameter sets. For the sake of clarity, the stopping point selection is outside of the scope of this paper and our intent is to show the strength of learning successful regularization from a lower dimensional pilot input \cite{ulyanov2018deep} for the analysis of the learning process. To show the general validity of obtained reconstructions we commit to the regularization network with parameters $\theta$ where the relative loss ratio difference below $10^{-3}$ between batch gradient iterations. {Implementation details are explained in Section~\ref{subsec:exp_setup}.}

\subsection{Experimental Setup}
\label{subsec:exp_setup}
 In this section we explain data specifications and performance evaluation criteria to assess the performance of the proposed method.

\noindent
\subsubsection{Data}
\label{subsec:Method-Data}
We acquired DCE-MR images using a work in progress (WIP) sequence from 10 patients (8 female, 2 male, age=6.5$\pm$4.9 years) who underwent a contrast-enhanced MRI exam \cite{coll2021modeling,coll2020bulk} for their clinical evaluation. All data were acquired following an approved IRB protocol and obtaining consent. After 20 seconds of imaging, a CA (gadobutrol, 0.75~ml/kg, Bayer HealthCare, Whippany, NJ) was power-injected intravenously. We used a stack-of-stars 3D FLASH prototype sequence with a multi-channel body-matrix coil (3T Siemens Skyra/Prisma/Trio, TR/TE/FA=3.56/1.39~ms/12$^{\circ}$, 32 coronal slices, voxel size=1.25x1.25x3~mm$^3$, 3978 radial spokes acquired in 6.24 minutes with golden angle radial ordering), average temporal resolution was 3.2~s/volume.
\noindent
\subsubsection{Performance Evaluation}
\label{subsec:Experiments_setup}
The proposed DP approach is compared with state-of-the-art temporal total-variation penalized CS solutions \cite{feng2014golden}. Instead of iteratively solving sparse temporal difference regularized minimization problem, our proposed learned regularization based image reconstruction approach updates parameters of the deep reconstruction map using the gradients from a squared error loss term between NUFFT of the output image and the observed k-space samples. Due to learned regularization posing a prior on the reconstruction scheme, we call our proposed approach learned deep-prior (DP) regularization. CS and the proposed DP are compared to each other in reconstructing sequences of DCE-MR images $X= \lbrace x_t | t\in \lbrace 1,2,\cdots,N \rbrace, x_t \in \mathbb{C}^{n\times n} \rbrace$ with $n = 224$. Each method reconstructs slices independently. 

To propose a fair comparison baseline with CS, different regularization parameter values $\lambda = \lbrace 0, 0.00125,$ $ 0.0125, 0.125, 1.25 \rbrace$ are presented. Observe that $\lambda =0$ directly yields the inverse-NUFFT solution. As the value of the regularization parameter increases, it is expected to obtain reconstructions that are smooth both in image and time domains. Both reconstruction methodologies utilize the independence of observed slices, reconstruct one slice a time and operate on k-space observations $k_t$s. For each image reconstruction $x_t$ at time point $t$, $k_t$ includes $sp=34$ spokes that corresponds to 3.3 seconds of acquisition (i.e., temporal resolution). Each method reconstructs more than 3[minutes] of acquisition for different scenarios that differ in patient, coil-setup acquisition device, etc. To summarize; for different patient abdomen reconstructions $k_t = \mathbb{C}^{224 \times c \times 34}$ where $c$ (number of coils) differs between patients for $t\in\lbrace 1 ,2,\cdots, N\rbrace $ with $N= 55$.

Both of these approaches were implemented with Python programming language. For NUFFT and its inverse Muckley's Pytorch Keiser-Bessel NUFFT implementation, v1.1 is used \cite{muckley_tkbnufft}. Each CS reconstruction ($\lambda \neq 0 $) performs 24 conjugate gradient descent iterations each iteration step size is determined with line search. DP is implemented using PyTorch v1.7.1 with Cuda v10.2. Both CS solution and DP training is performed on a workstation with Quadro-P5000 GPU and Intel-Xeon E5-2698 v4 @ 2.20GHz CPU. Model parameters are initialized with random weights. Weights are updated with a batch gradient descent update. Each batch includes $|B|= 16$ time points ($= 16 \times 34$ spokes) that contribute to the gradient and DP is trained for at least 2000 epochs with a learning rate of $10^{-4}$. Source code for DP method is publicly available \footnote{{https://github.com/quin-med-harvard-edu/kidney-LR-DCEMRI/}}.
\section{Results}
\label{sec:Experiments}
This section describes the results of a detailed set of experiments described under Section~\ref{subsec:Experiments_setup} with the data explained under Section~\ref{subsec:Method-Data}. 
We investigate the differences between CS and DP reconstruction using two different sets of experiments, one for assessing visual quality, second for assessing the accuracy of the estimated quantitative parameter. In Section~\ref{subsec:Experiments_vis} image domain smoothness and temporal structure analysis is provided. Section~\ref{subsec:Experiments_param} shows DP produces consistent filtration rates. 
{We encourage reader to the electronic copy of this document for higher resolution images.}

\subsection{Visual Reconstruction Quality}
\label{subsec:Experiments_vis}

%
\begin{figure}[t]
\centering
\begin{minipage}{0.70\columnwidth}
		\centering
		\includegraphics[width=\columnwidth]{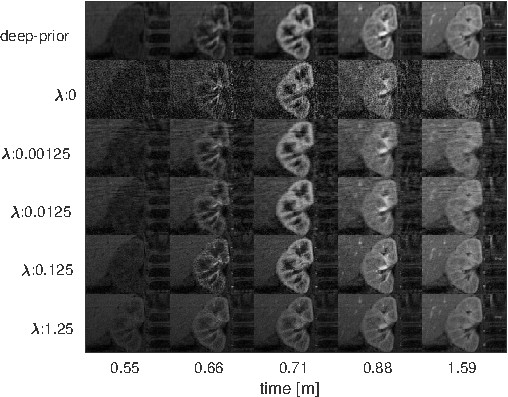}
\end{minipage} \\
\vspace{0.2cm}
\begin{minipage}{0.70\columnwidth}
		\centering
		\includegraphics[width=\columnwidth]{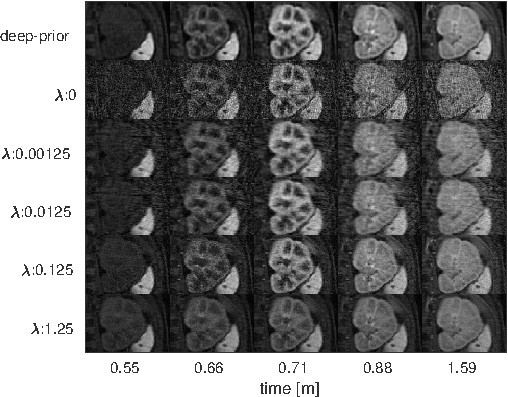}
\end{minipage}
      \caption{
      \label{fig:matrix_fig}
      Reconstructed images are shown in detail for the left and right kidney regions for patient-6. Each sub-figure employs a matrix form where each row represents a different method of reconstruction and each row refers to a different time point. Time stamps visualize a no-contrast phase, a cortical phase, and two medullary phases respectively. Images are plotted after applying a normalization by matching the respective baseline images. We utilized the first time-frame ($t=0.55[m]$) to do baseline normalization for each respective method. First row represents DP reconstruction where the following rows are generated using CS with $\lambda = [0,0.00125,0.0125,0.125,1.25 ]$ in order. Observe that increasing $\lambda$ eliminates reconstruction noise but also oversmooths the temporal signal and degrades the contrast differences. Distinctly, the proposed DP produces high quality images, with low in-image noise without degrading the temporal and spatial contrast differences.
      }
\end{figure}
%
%
\begin{figure*}[t]
\centering
\begin{minipage}{0.36\columnwidth}
		\centering
		\includegraphics[width=.95\columnwidth]{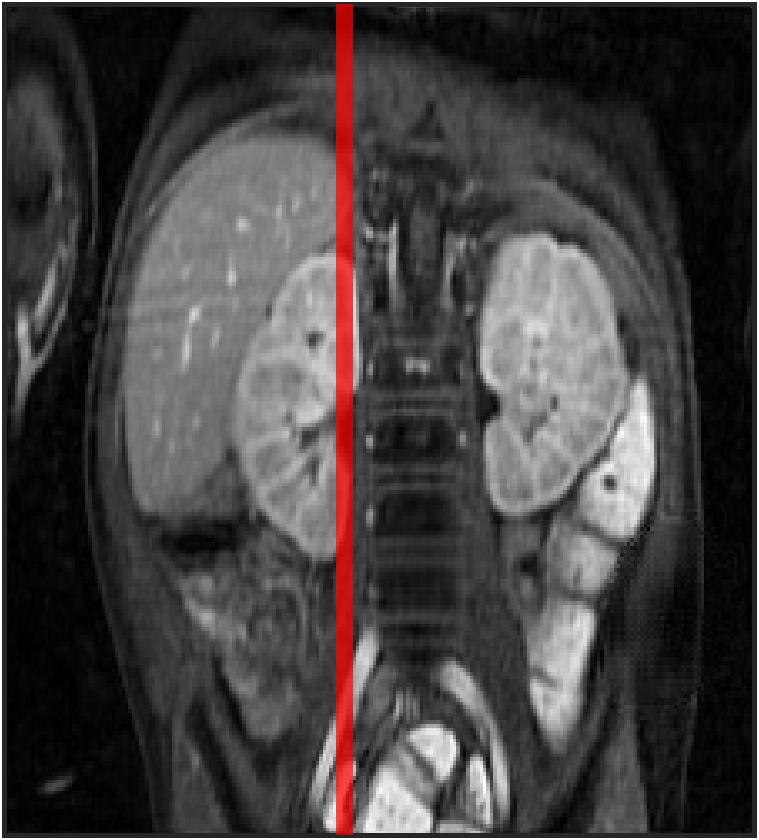}
\end{minipage}
\begin{minipage}{0.65\columnwidth}
		\centering
	   \subfloat{\includegraphics[width=.48\columnwidth]{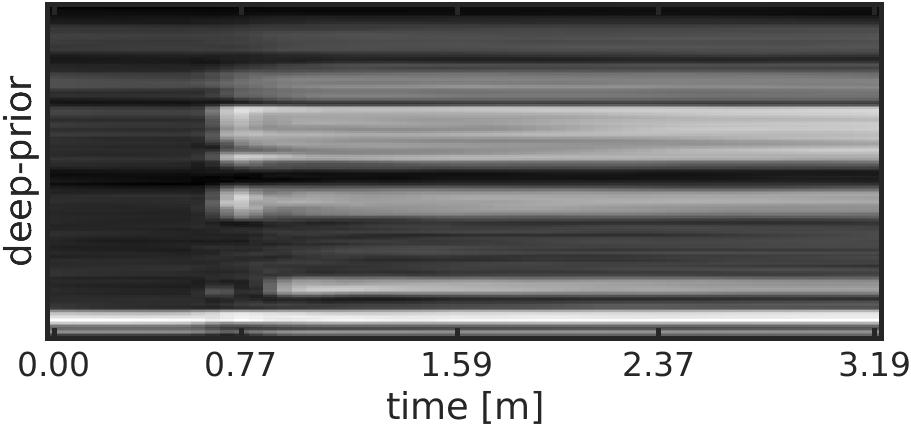}}
	   \subfloat{\includegraphics[width=.48\columnwidth]{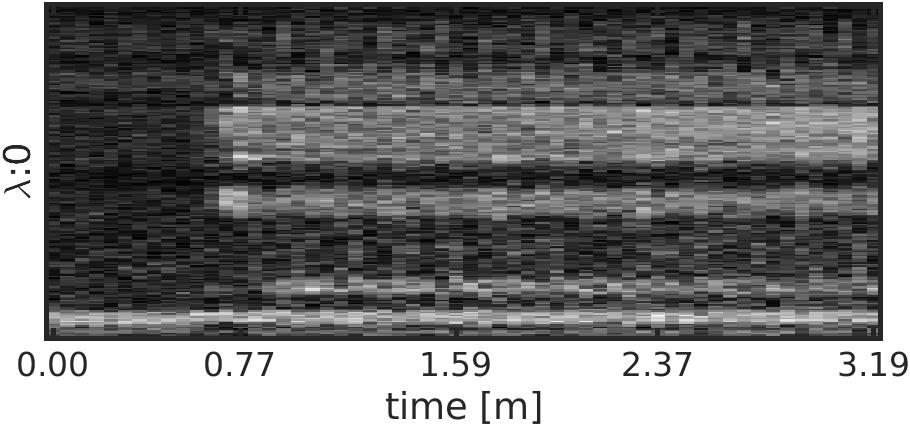}}\\ \vspace{-0.2cm}
       \subfloat{\includegraphics[width=.48\columnwidth]{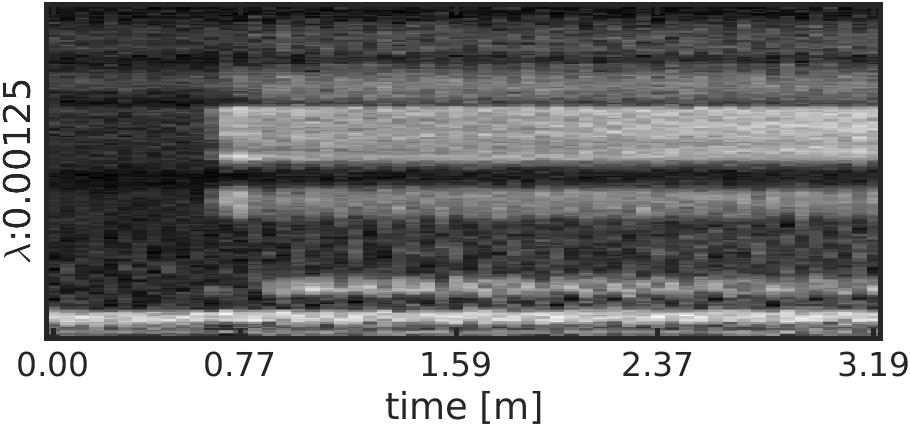}}
       \subfloat{\includegraphics[width=.48\columnwidth]{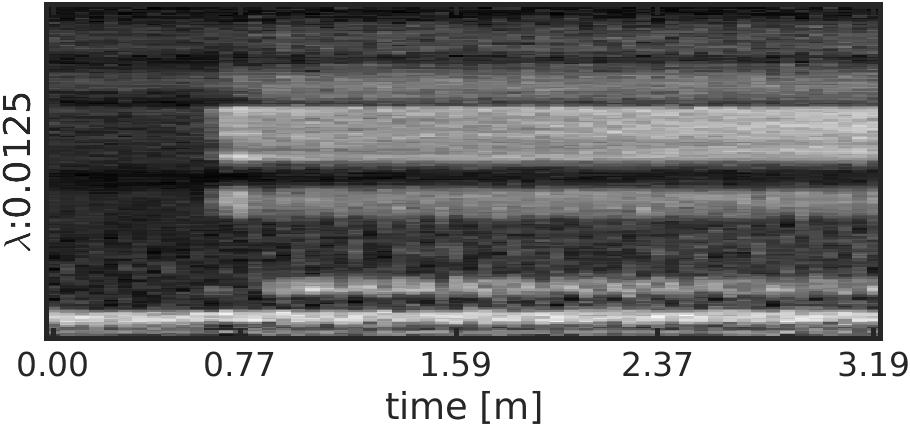}}\\ \vspace{-0.2cm}
       \subfloat{\includegraphics[width=.48\columnwidth]{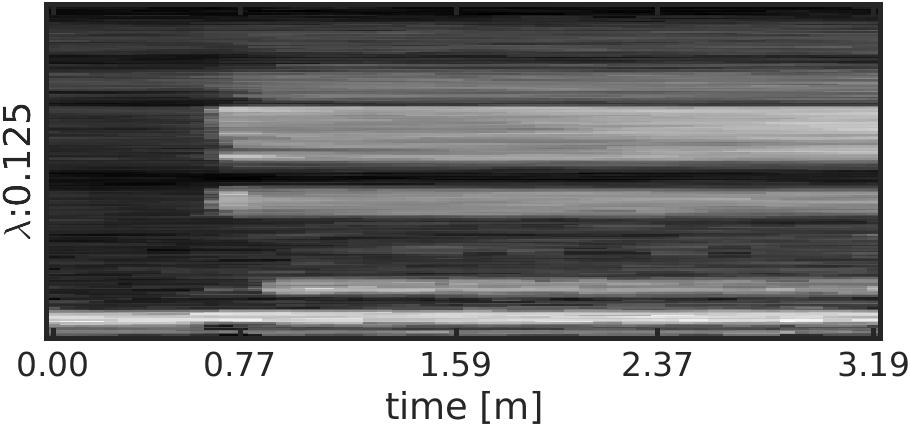}}
	   \subfloat{\includegraphics[width=.48\columnwidth]{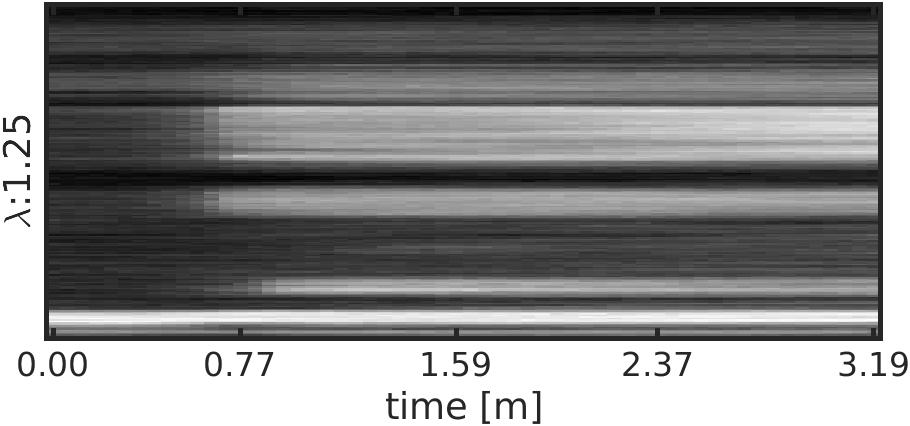}}\\ \vspace{-0.1cm}
\end{minipage}\vline
\begin{minipage}{0.36\columnwidth}
		\centering
		\includegraphics[width=.95\columnwidth]{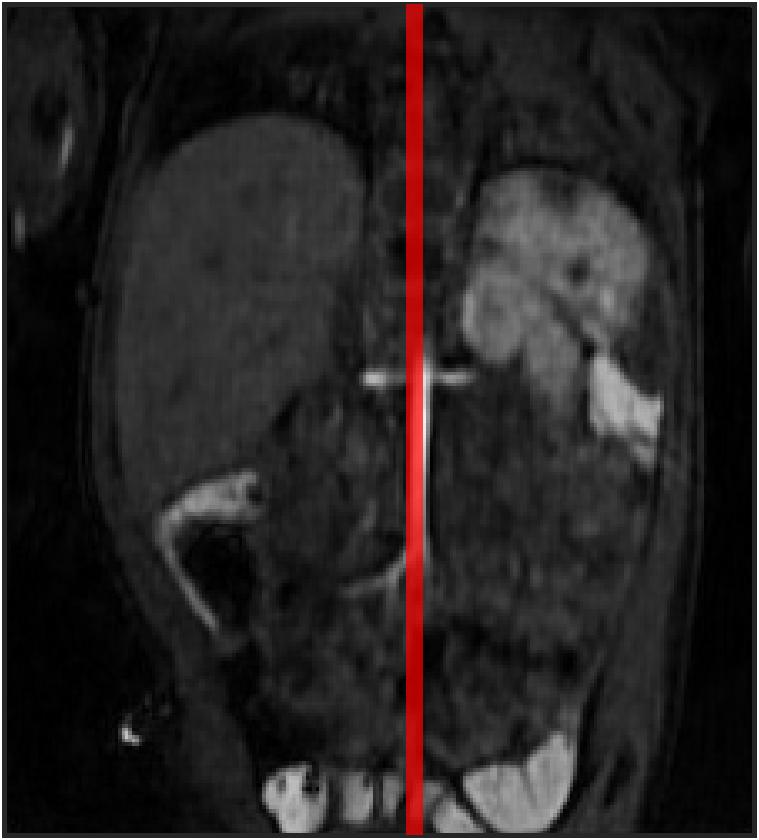}
\end{minipage}
\begin{minipage}{0.65\columnwidth}
		\centering
	   \subfloat{\includegraphics[width=.48\columnwidth]{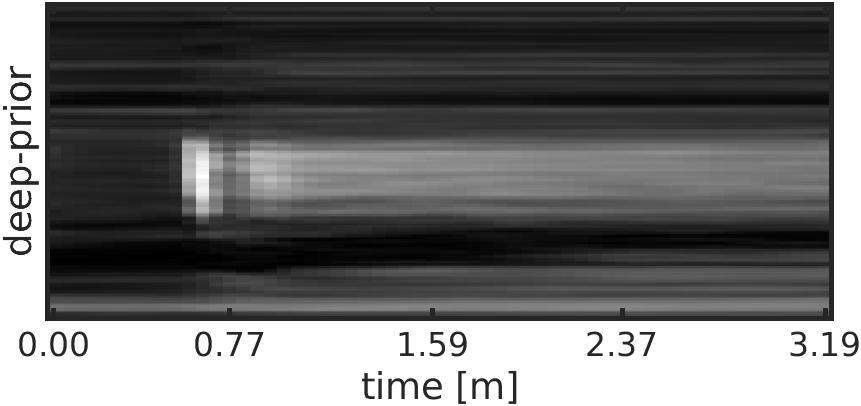}}
	   \subfloat{\includegraphics[width=.48\columnwidth]{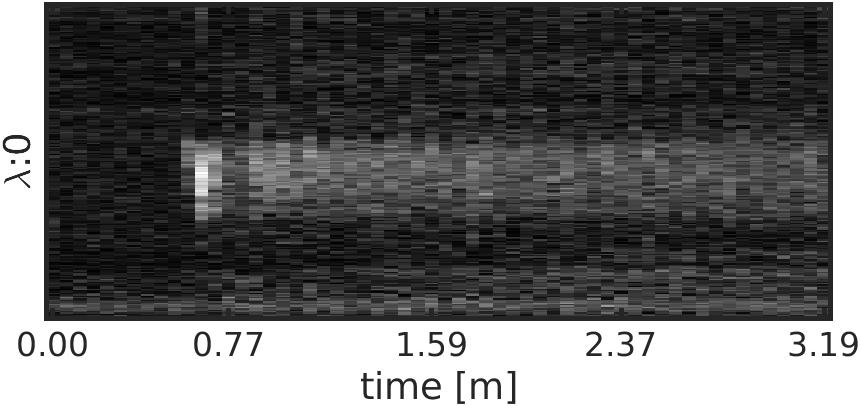}}\\ \vspace{-0.2cm}
       \subfloat{\includegraphics[width=.48\columnwidth]{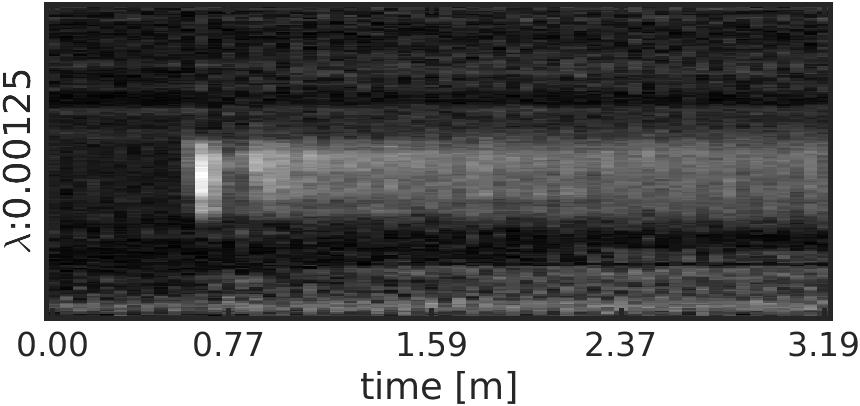}}
       \subfloat{\includegraphics[width=.48\columnwidth]{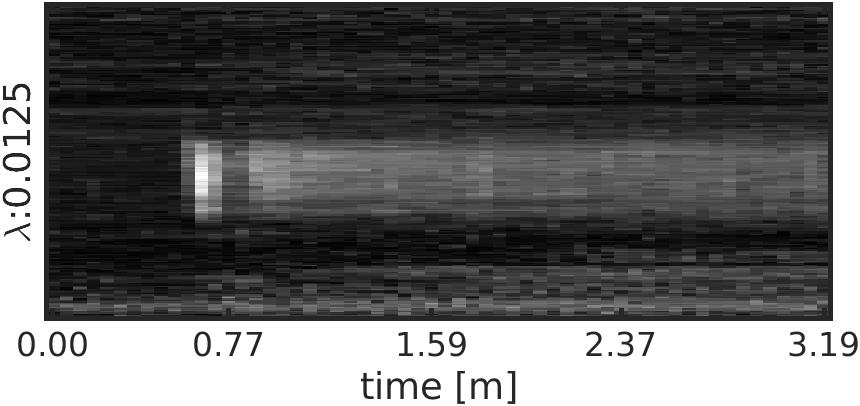}}\\ \vspace{-0.2cm}
       \subfloat{\includegraphics[width=.48\columnwidth]{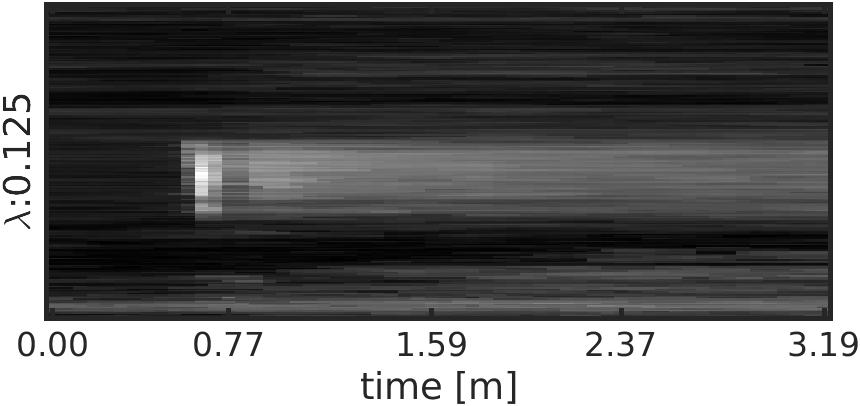}}
	   \subfloat{\includegraphics[width=.48\columnwidth]{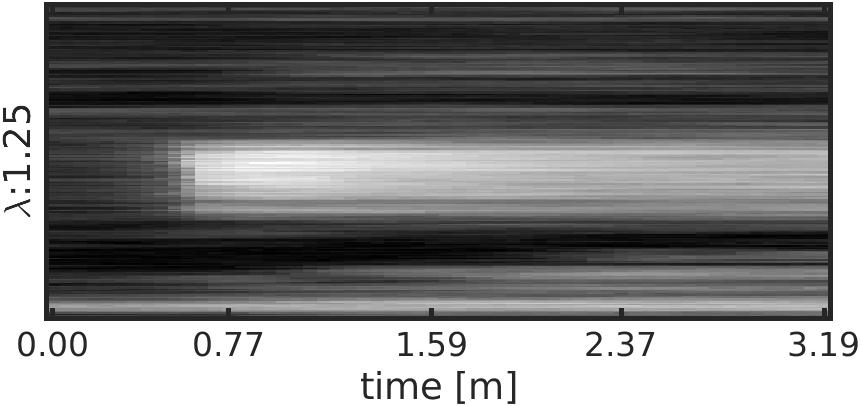}}\\ \vspace{-0.2cm}
\end{minipage}
    \footnotesize
    \caption{\label{fig:cut_fig_sub_3}
      Left panel shows kidney and aorta slices reconstructed by the proposed DP algorithm for patient-2 at a time instance $t$. The  red line represents a vertical column of voxels selected from the image for which we plot temporal signal change for each reconstruction method. Right panel shows the temporal signal chanhe plot of the selected vertical image column for DP (deep prior) and CS techniques with different regularization parameters ($\lambda$). These time plots are shown for kidneys in the top panel, and for aorta  in the bottom panel.  Observe that, increasing $\lambda$ creates smoother temporal signal, causing the loss of temporal structure of the signal. Specifically, in the aorta slice (bottom panel), it is clearly seen that high $\lambda$ values completely smooth out the expected second contrast peak. }
\end{figure*}
\begin{table*}[t]
\caption{{This table summarizes mean and standard deviation for each method on each patient respectively (excluding patient-8 which is already represented. \textit{DP} row number is determined based on the averaged TV value over all patients where top row indicates the highest average TV. It is possible to observe that DP removes noise on background image resulting in TV terms even lower than $\lambda=1.25$ on average. Additionally, DP preserves the temporal differences on contrast segment and avoids over smoothing. Therefore, DP combines the strengths of different hyperparameter values yielding a desired reconstruction.}
        \label{tab:tv_table} \vspace{-0.3cm}}
{\def\arraystretch{0.3}\tabcolsep=4pt
    \begin{center}
        \footnotesize
        \begin{tabular}{c c| c c c c c c c c c}
            & &\multicolumn{9}{c}{Total Variation (mean$\pm$std)}\\
            &method &p-1 &p-2 &p-3 &p-4 &p-5 &p-6 &p-7 &p-9 &p-10 \\
            \midrule
            \parbox[t]{2mm}{\multirow{6}{*}{\rotatebox[origin=c]{90}{contrast}}}
            &$\lambda:$0&  0.22 $\pm$   0.47&  0.40 $\pm$   0.63&  0.37 $\pm$   0.61&  0.26 $\pm$   0.51&  0.58 $\pm$   0.76&  0.53 $\pm$   0.73&  0.61 $\pm$   0.78&  0.82 $\pm$   0.90&  0.16 $\pm$   0.40\\ 
            &\textbf{DP} &  0.28 $\pm$   0.53&  0.21 $\pm$   0.46&  0.24 $\pm$   0.49&  0.23 $\pm$   0.48&  0.24 $\pm$   0.49&  0.25 $\pm$   0.50&  0.25 $\pm$   0.50&  0.19 $\pm$   0.43&  0.09 $\pm$   0.30\\
            &$\lambda:$0.00125 &  0.11 $\pm$   0.34&  0.18 $\pm$   0.43&  0.12 $\pm$   0.34&  0.14 $\pm$   0.38&  0.17 $\pm$   0.41&  0.07 $\pm$   0.26&  0.27 $\pm$   0.52&  0.20 $\pm$   0.45&  0.04 $\pm$   0.19\\
            &$\lambda:$0.0125&  0.07 $\pm$   0.27&  0.14 $\pm$   0.38&  0.07 $\pm$   0.27&  0.14 $\pm$   0.38&  0.14 $\pm$   0.37&  0.02 $\pm$   0.16&  0.18 $\pm$   0.43&  0.21 $\pm$   0.46&  0.02 $\pm$   0.15\\ 
            &$\lambda:$0.125&  0.00 $\pm$   0.04&  0.02 $\pm$   0.13&  0.01 $\pm$   0.12&  0.03 $\pm$   0.17&  0.02 $\pm$   0.13&  0.03 $\pm$   0.16&  0.02 $\pm$   0.13&  0.10 $\pm$   0.32&  0.01 $\pm$   0.12\\
            &$\lambda:$1.25&  0.00 $\pm$   0.04&  0.01 $\pm$   0.08&  0.00 $\pm$   0.06&  0.02 $\pm$   0.12&  0.01 $\pm$   0.08&  0.01 $\pm$   0.11&  0.00 $\pm$   0.07&  0.02 $\pm$   0.15&  0.01 $\pm$   0.08\\
            \midrule 
            \parbox[t]{2mm}{\multirow{6}{*}{\rotatebox[origin=c]{90}{bckgrnd}}}
            &$\lambda:$0&  2.92 $\pm$   1.71&  3.96 $\pm$   1.99&  5.53 $\pm$   2.35&  4.66 $\pm$   2.16&  3.30 $\pm$   1.82&  2.92 $\pm$   1.71&  5.57 $\pm$   2.36&  1.69 $\pm$   1.30&  3.86 $\pm$   1.97\\
            &$\lambda:$0.00125&  2.21 $\pm$   1.49&  2.77 $\pm$   1.66&  3.79 $\pm$   1.95&  4.26 $\pm$   2.06&  2.40 $\pm$   1.55&  1.88 $\pm$   1.37&  4.11 $\pm$   2.03&  1.75 $\pm$   1.32&  2.75 $\pm$   1.66\\  
            &$\lambda:$0.0125&  1.26 $\pm$   1.12&  1.70 $\pm$   1.30&  3.14 $\pm$   1.77&  3.44 $\pm$   1.85&  2.00 $\pm$   1.42&  1.07 $\pm$   1.04&  3.01 $\pm$   1.74&  1.51 $\pm$   1.23&  1.91 $\pm$   1.38\\ 
            &$\lambda:$0.125&  0.35 $\pm$   0.59&  0.59 $\pm$   0.77&  1.18 $\pm$   1.09&  1.21 $\pm$   1.10&  0.37 $\pm$   0.61&  0.42 $\pm$   0.65&  1.30 $\pm$   1.14&  0.33 $\pm$   0.57&  0.64 $\pm$   0.80\\ 
            &$\lambda:$1.25&  0.31 $\pm$   0.56&  0.49 $\pm$   0.70&  0.63 $\pm$   0.80&  1.25 $\pm$   1.12&  0.47 $\pm$   0.68&  0.31 $\pm$   0.56&  0.69 $\pm$   0.83&  0.25 $\pm$   0.50&  0.53 $\pm$   0.73\\
            &\textbf{DP} &  0.27 $\pm$   0.52&  0.28 $\pm$   0.53&  0.28 $\pm$   0.53&  0.27 $\pm$   0.52&  0.29 $\pm$   0.54&  0.28 $\pm$   0.53&  0.28 $\pm$   0.53&  0.19 $\pm$   0.44&  0.53 $\pm$   0.73\\  
        \end{tabular}
    \end{center}
    }
    
\end{table*}
%
\begin{figure}[t]
\centering
\begin{minipage}{0.35\columnwidth}
		\centering
		\includegraphics[width=.95\columnwidth]{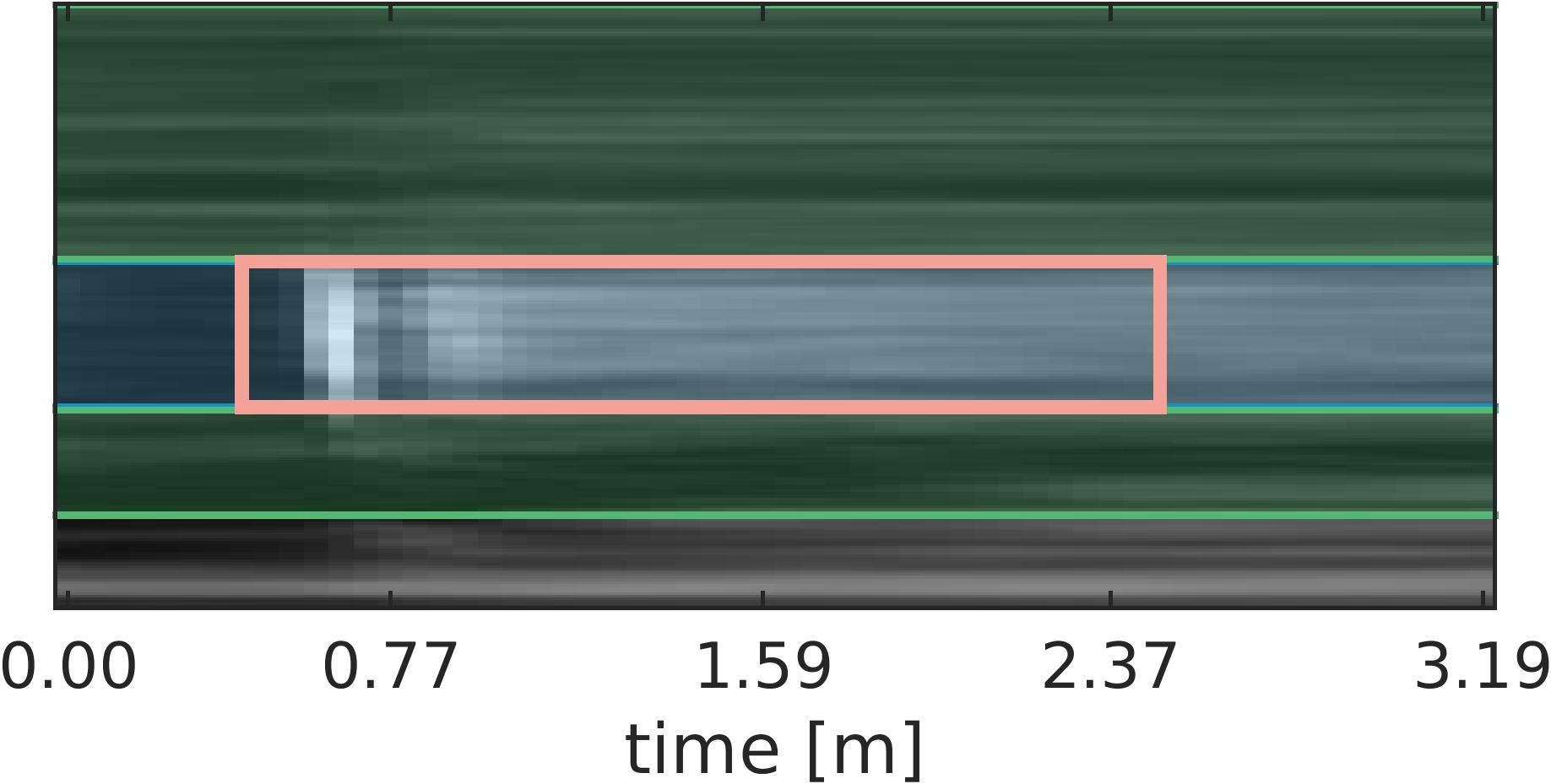}
\end{minipage}
\begin{minipage}{0.52\columnwidth}
		\centering
		\includegraphics[width=.95\columnwidth]{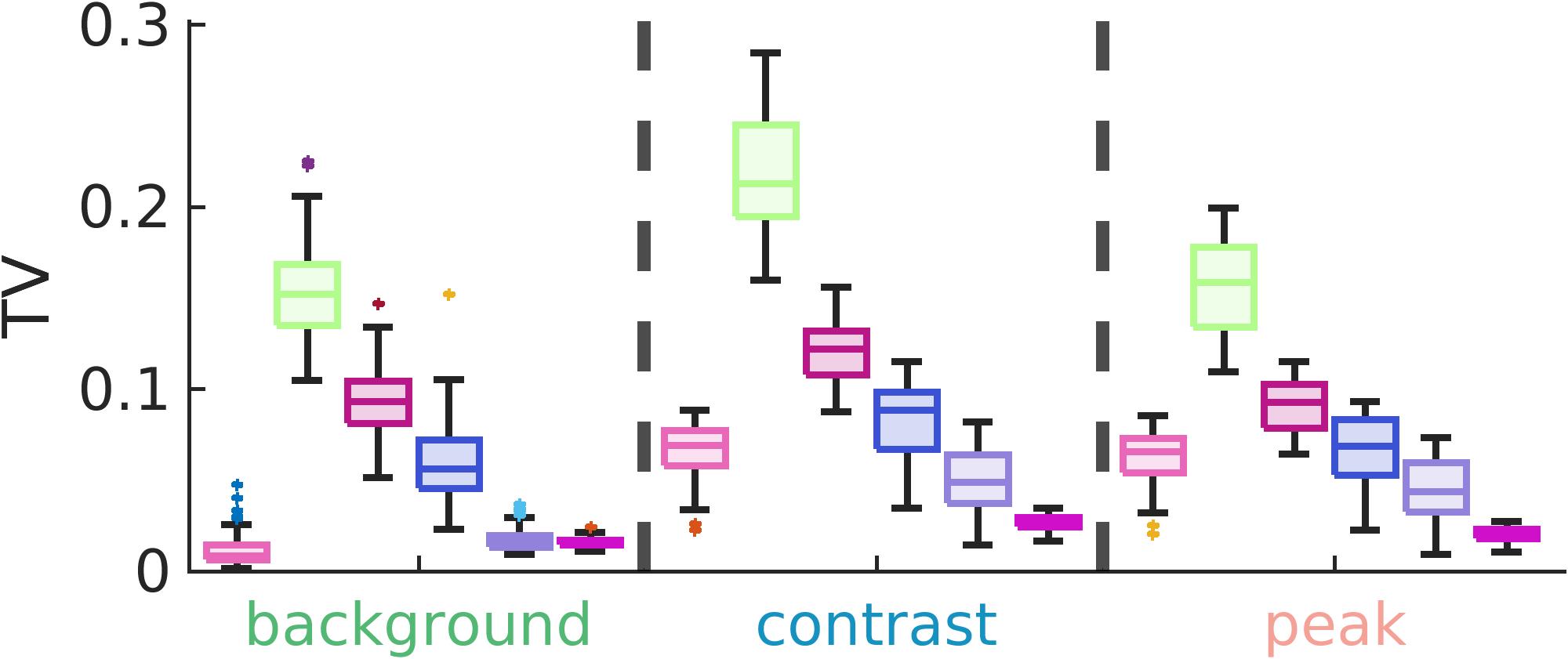}
\end{minipage}
\begin{minipage}{0.1\columnwidth}
		\centering
		\includegraphics[width=.95\columnwidth]{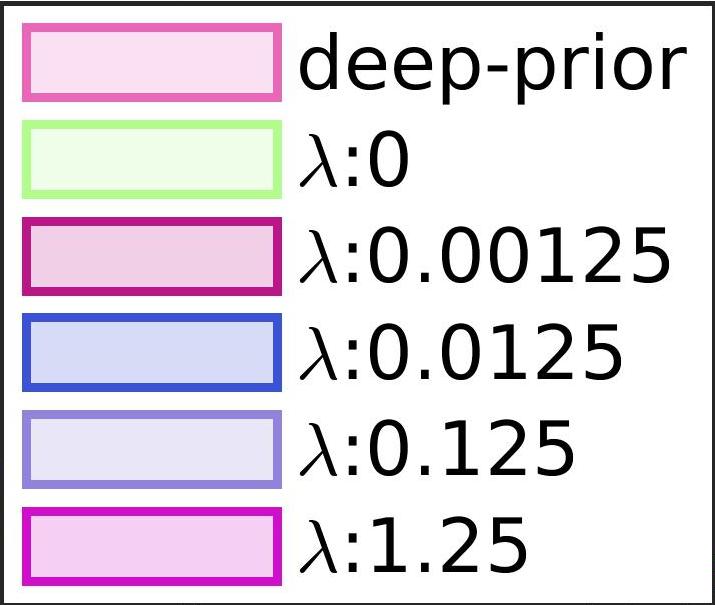}
\end{minipage}
      \caption{\label{fig:box_fig_sub_8}
      Temporal total variation (TV) is computed and  compared for different methods in a sample patient. An arterial slice-cut is plotted to show the temporal signal. This data is then segmented into the peak contrast, contrast and background sections that are expected to have different temporal TV values. The proposed deep-prior achieved lowed temporal TV in the background regions suppressing noise but preserved the temporal dynamics of the signal achieving similar temporal TV to the one from a CS reconstruction with a lower $\lambda$ value.
      }
\end{figure}

In CS reconstruction increasing the regularization parameter $\lambda$ value utilizes the piece wise constancy of the image sequence through time. This assumption reduces the noise presence utilizing the nature of the image especially around areas without contrast agent (CA). However, this assumption staggers in the presence of CA, especially in regions such as aorta, where fast signal changes are observed. This section discusses the benefits of DP approach in four sets of experiments. 

{CA differences between phases (temporal intensity differences) is crucial for parametric estimation of essential physiological parameters. First set of experiments focus on illustrating}
DP preserving CA differences while suppressing the noise. To emphasize the overall smoothing effect on an interest area of the reconstruction,
Figure~\ref{fig:matrix_fig} illustrates a sequence of time-stamped images from a patient. This figure specifically focuses on illustrating the details of the left and right kidneys, and their enhancement patterns separately. Time stamps visualize a no-contrast phase, a cortical phase and two medullary phases, respectively. Observe that DP ($1^\text{st}$ row) eliminated reconstruction noise in CA to lower $\lambda$ CS reconstructions (e.g., $2^\text{nd}, \ 3^\text{rd}$ rows). On the other hand, increasing regularization parameter $\lambda$, as expected further reduces noise at the cost of temporal over-smoothing. 
It is possible to observe from $3^\text{rd}$ and $4^\text{th}$ columns of each sub-image that increasing $\lambda: 0 \rightarrow 1.25$ yields over-smoothing through time and ultimately starts to remove the CA enhancement effect for large $\lambda$s. {Particularly CS with $\lambda = 1.25$ (as the last row in both sub images) resulted in marginal intensity differences.}

{ Similarly, preserving intensity differences on cross-sections for different interest voxels (e.g. aorta) are also important for parametric estimation.}
Figure~\ref{fig:cut_fig_sub_3} shows two image slices,  one including kidneys and one including aorta for a patient which illustrates the time behavior of CS and DP reconstructions.  In this figure the first column shows the DP reconstruction of the corresponding slice with a vertical cross section highlighted with a red line for a given $\text{idx}$. Second column is chunked into different voxel-time plots through the respective cross-section that is highlighted in the first column figure. Illustrated methods in the second column are DP, i-NUFFT and CS for $\lambda \in \lbrace 0.00125, 0.0125, 0.125, 1.25 \rbrace$ respectively. Kidney cross section is selected specifically to cover the behavior of the CA on kidney and the artery next to the bladder. As observed from the time figures, increasing $\lambda$ decreases the time domain noise on the cross section. However, in return it over-smooths the curve on the kidney reconstruction bringing baseline value and peak CA value closer to each other. As can be observed from the image, specifically $\lambda \geq 0.125$ reconstruction time illustrations, the baseline value and the peak value are closer in pixel value and moreover there exist a smooth transition on the artery next to the bladder (bottom pixel values) which contradicts with expected physiological response. On the other hand, DP retains all physiological structure of the data like lower $\lambda$ values without carrying over the noise. Similarly increasing the temporal piece-wise constancy term removes the noise and in return disfavors the estimation structure of the arterial slice. Specifically for aorta slice reconstruction $\lambda = 1.25$ removes the second peak of the arterial curve. On the other hand, DP reconstruction retains the double peak structure of arterial curve without any further noise.

{ In this set of experiments we demonstrate that unlike CS approaches, DP successfully reduces noise on background voxels on dynamic images without the loss of temporal intensity differences on CA effected voxels. Temporal total variation (TV) is an indicative measure of temporal differences in the signal (see Section~\ref{subsubsec:cs_methods}).  Preserving temporal intensity differences on CA effected interest areas result in higher TV values whereas for background pixels TV is expected to be low.}
Following Figure~\ref{fig:cut_fig_sub_3}, an arterial slice-cut is selected to represent temporal relation for each patient. As illustrated in Figure~\ref{fig:box_fig_sub_8} for a sample patient, this data is further segmented into the contrast and background sections that are expected to have different TV values. Observe that DP attains lower TV values on the background while relaxing the penalization for the CA effected pixel rows. Therefore, in contrast to CS, DP reduces the noise effects spatially enforcing piecewise constancy through time for compartments that are not CA effected while preserving the CA effect where needed. Table~\ref{tab:tv_table} summarizes mean and standard deviation of TV values for each method on each patient respectively (excluding patient-6 which is already represented in first row). Observe that on average, for the CA enhanced tissue regions, DP attains temporal TV values that are close to lower $\lambda$ values while suppressing the noise on the background regions.

%
\begin{figure}[t]
\centering
\begin{minipage}{0.22\columnwidth}
		\centering
		\includegraphics[width=\columnwidth]{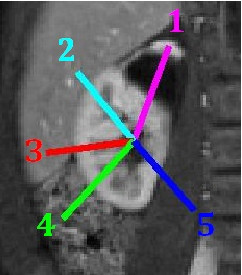}
\end{minipage}
\vline
\begin{minipage}{0.76\columnwidth}
		\centering
	   \subfloat{\includegraphics[width=.31\columnwidth]{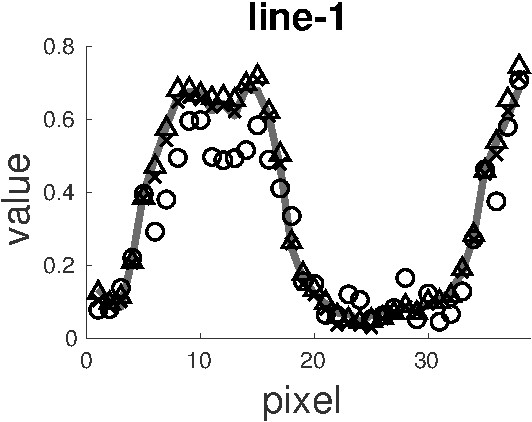}}
	   \subfloat{\includegraphics[width=.31\columnwidth]{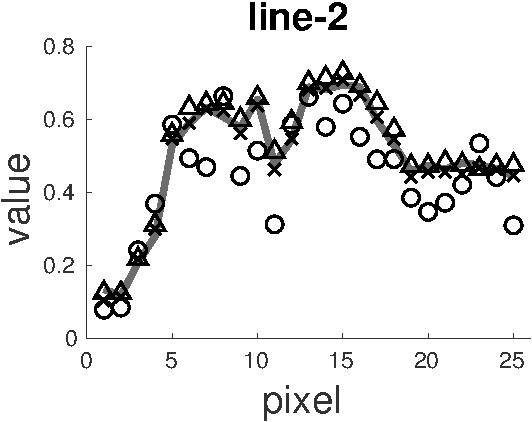}}
       \subfloat{\includegraphics[width=.31\columnwidth]{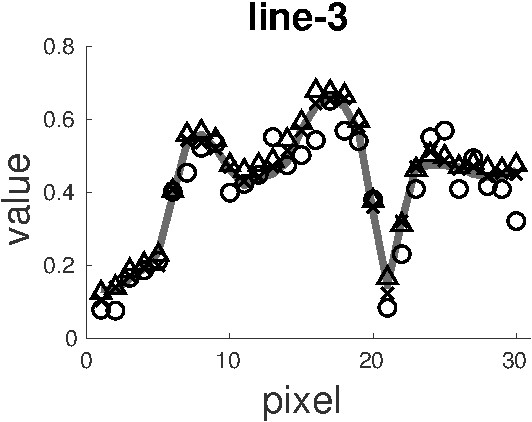}}\\ \vspace{-0.2cm}
       \subfloat{\includegraphics[width=.31\columnwidth]{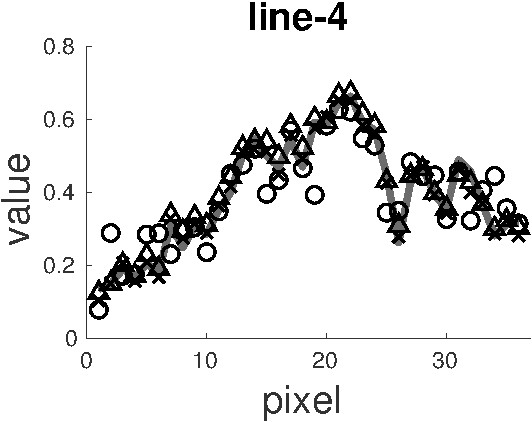}}
       \subfloat{\includegraphics[width=.31\columnwidth]{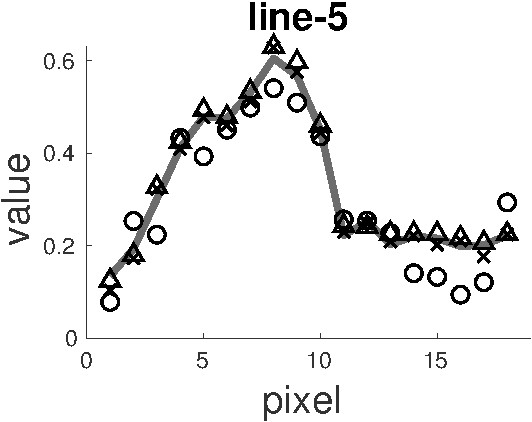}}
       \subfloat{\includegraphics[width=.31\columnwidth]{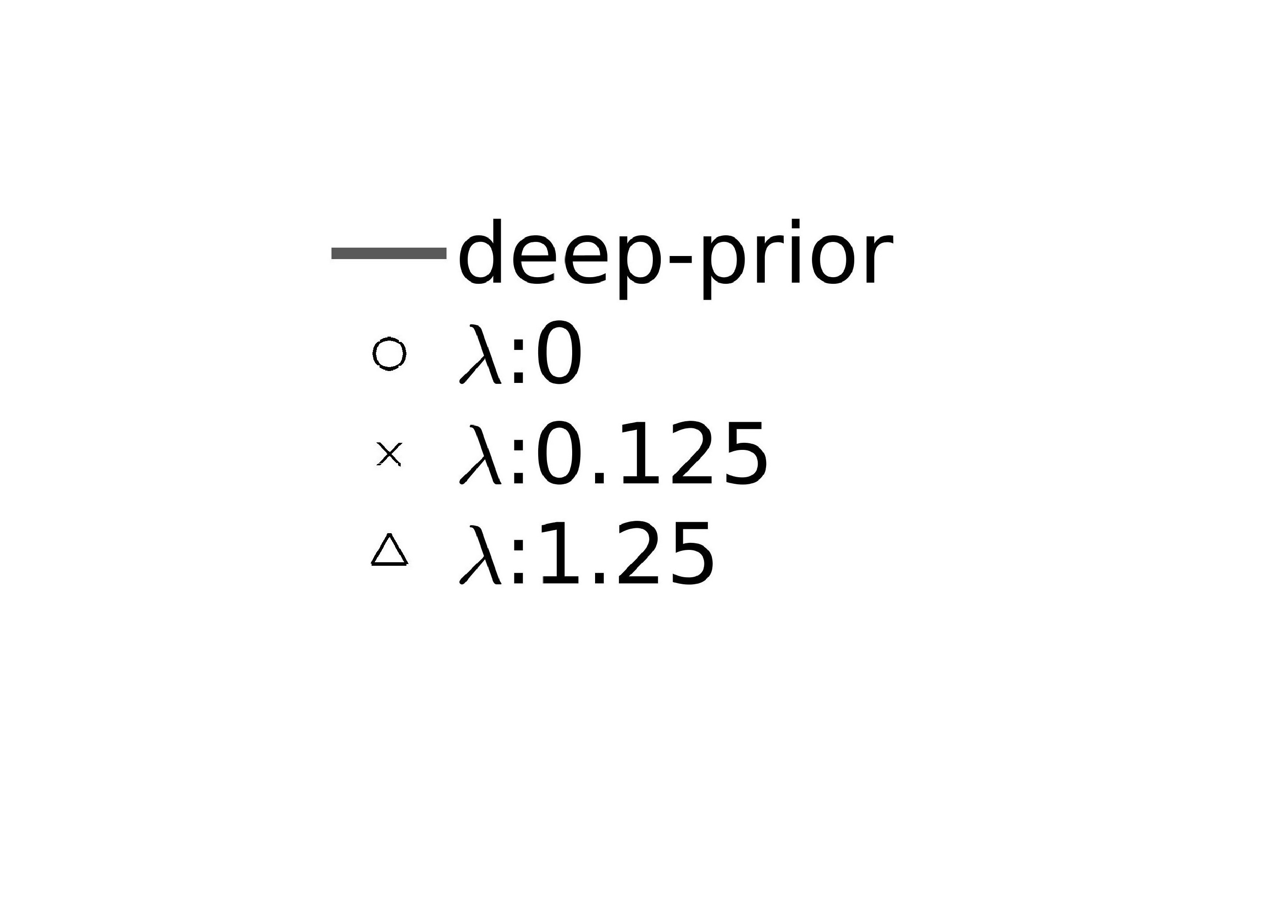}}
        \vspace{-0.2cm}
\end{minipage}
    \footnotesize
    \caption{\label{fig:line_fig}
    Pixel values along different line segments along the kidneys are
shown. Numerated line segments are illustrated on a DP reconstructed image
(on the left). Pixel values on each line segment are extracted after each respective
reconstruction is normalized. Plots on the right-hand side represent
in-image normalized pixel values for DP and CS with $\lambda \in \lbrace0, 0.125, 1.25\rbrace$
respectively. $\lambda$ $0.00125$ and $0.0125$ are omitted because of similar values
to $\lambda =0 $ and for neat presentation. As observed, DP provides comparable
spatial smoothness and edge strength to high $\lambda$ CS reconstructions, whereas
INUFFT reconstruction ($\lambda=0$) has noisy pixel values.
    }
\end{figure}
{ Streaking artifacts reduce the visual quality as we expressed earlier (see Figure~\ref{fig:matrix_fig}-row 2) and the main outcome of these artifacts are spatial intensity fluctuations in small neighborhoods. Therefore, we assess spatial noise effects with smoothness of spatial pixel intensity differences.} 
Figure~\ref{fig:line_fig} illustrates spatial smoothness of different reconstruction methodologies. This figure contains two compartments; The left sub-figure is the left-kidney DP reconstruction of patient-2 which also includes numerated line segments that on which the pixel values are extracted to show the smoothness of the image, the right sub figure contains 5 images that represent pixel values from the corresponding line segments from respective reconstructions. This figure intentionally focuses on 3 CS reconstructions excluding $\lambda$ values $0.00125$ and $0.0125$ as the smoothness values were similar to $\lambda =0 $. This also helps the readability of the figure. As shown in the figure, DP offers high $\lambda$ smoothness factors along different trajectories where $\lambda = 0$ results fluctuating values which are ultimately a reflection of the  remaining noise in the recontructed image. For example, due to reconstruction noise specifically in lines-(1,2) reconstruction with $\lambda=0$ provides lower pixel intensity values for high intensity areas. Additionally, on line-(5) the latter pixels share the intensity value for which $\lambda=0 $ yields variable intensity values. In all these examples DP provides smooth reconstructions matching $\lambda \in \lbrace 0.125 , 1.25\rbrace$.

\subsection{Tracer Kinetic Model Parameter Estimation}
\label{subsec:Experiments_param}
%
\begin{figure}[t]
\centering

\begin{minipage}{0.95\columnwidth}
		\centering
	   \subfloat{\includegraphics[width=0.32\columnwidth]{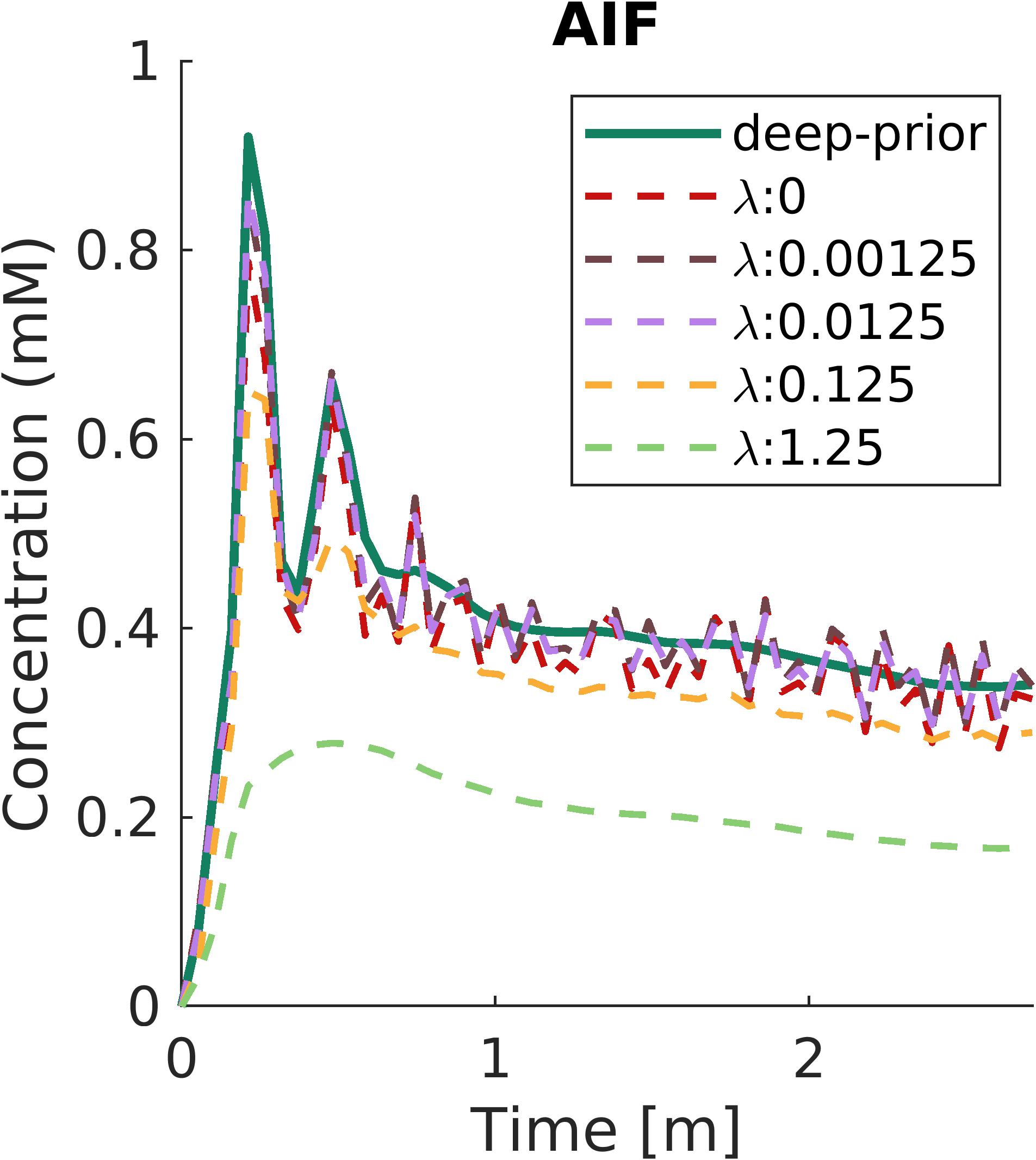}}\hfill
	   \subfloat{\includegraphics[width=0.32\columnwidth]{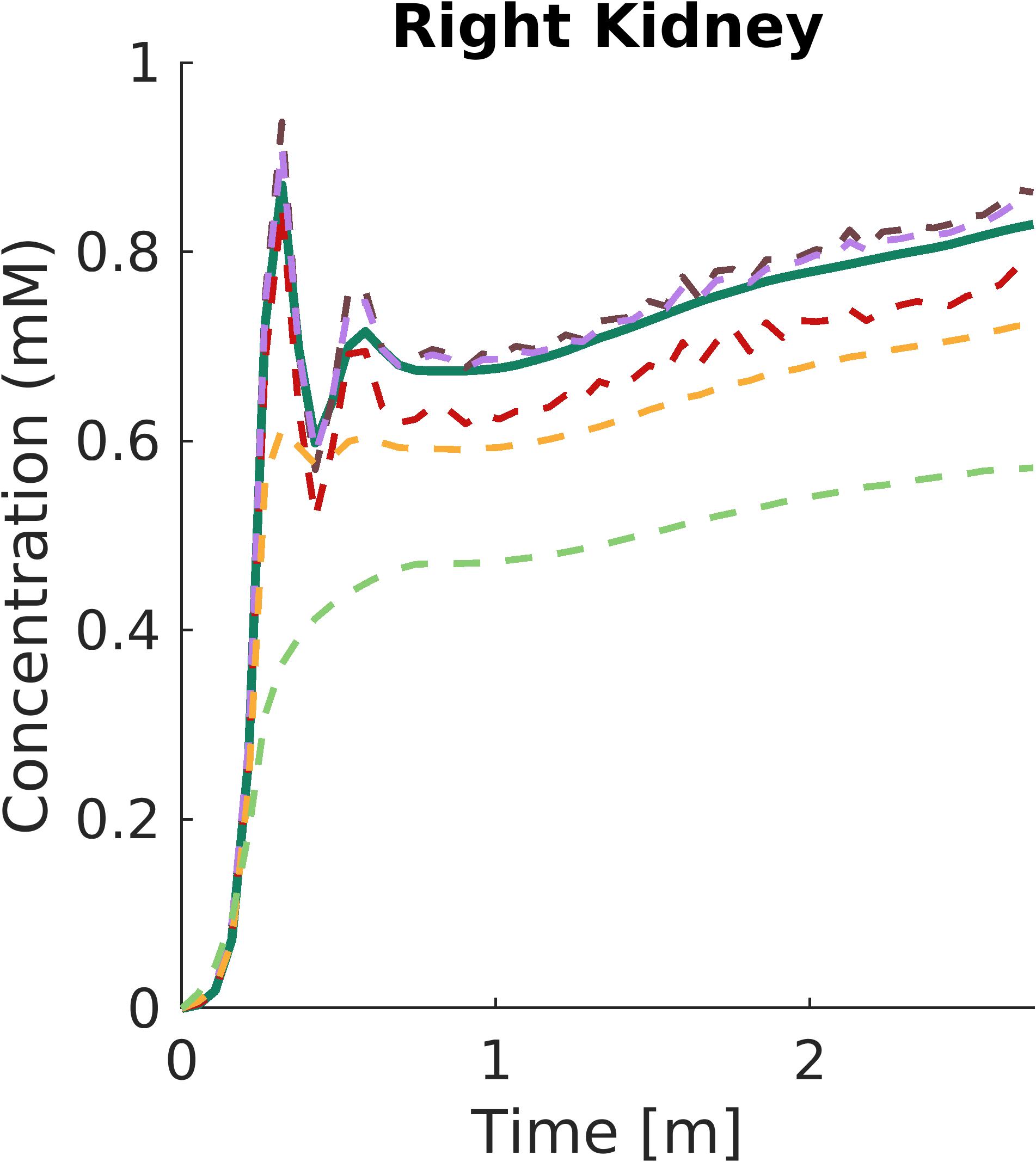}}\hfill
	   \subfloat{\includegraphics[width=0.32\columnwidth]{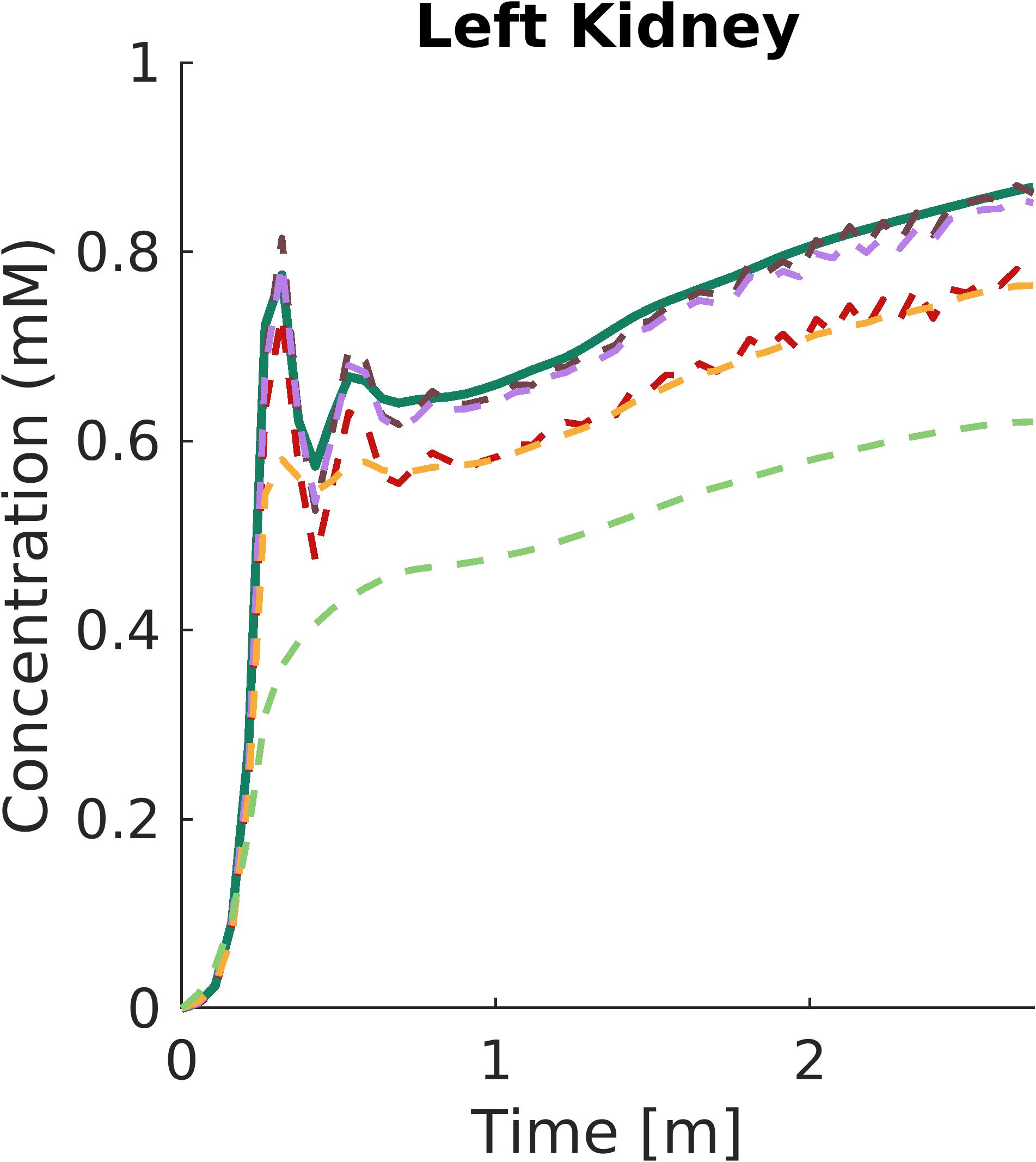}}\\
	   \vspace{-0.2cm}
	   \hspace{0.5cm}\textsf{(a)}\hspace{7.5cm}\\
	   \subfloat{\includegraphics[width=0.85\columnwidth]{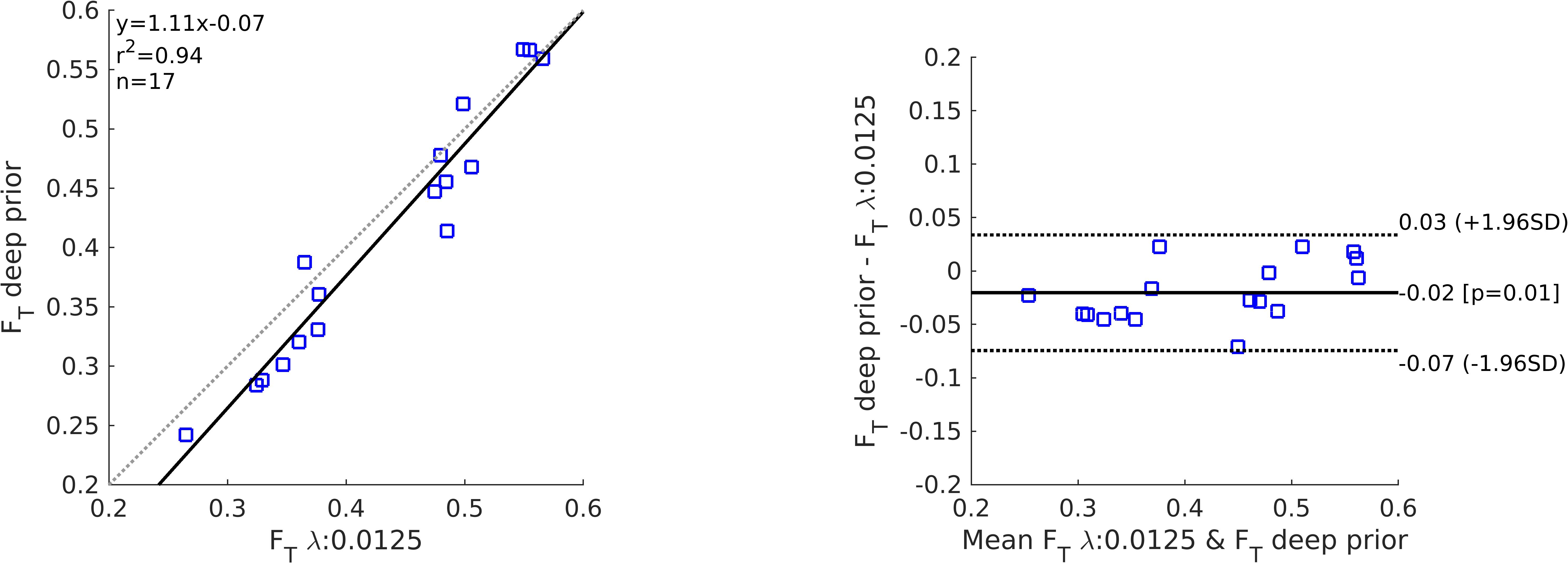}}\\
       \vspace{-0.2cm}
	   \hspace{0.5cm}\textsf{(b)}\hspace{7.5cm}\\
	   \subfloat{\includegraphics[width=0.85\columnwidth]{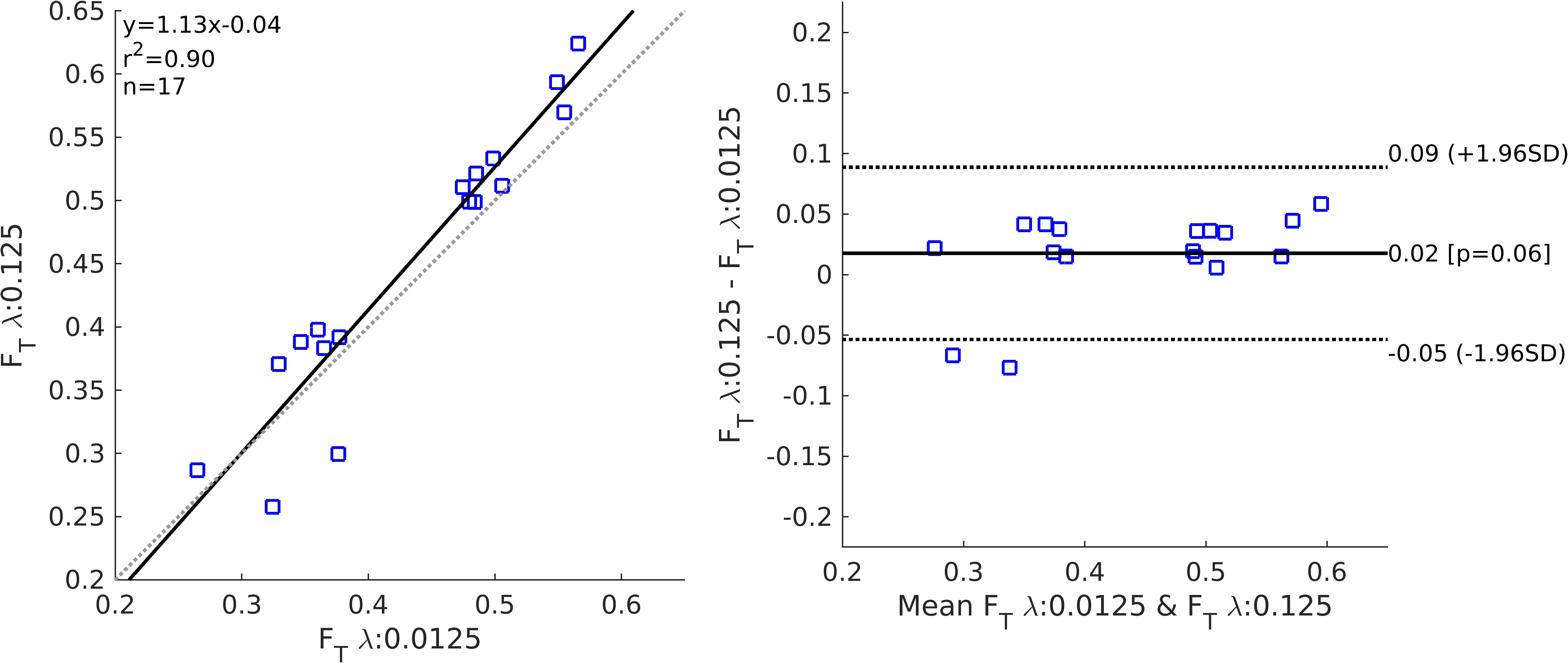}}\\
        \vspace{-0.2cm}
	   \hspace{0.5cm}\textsf{(c)}\hspace{7.5cm}\\
\end{minipage}
\\
\caption{\label{fig:curveFit_fig}
    For DP and all CS reconstructions, MRI signals averaged over aorta and kidney regions are converted into concentration curves and fitted to a 2-compartment tracer kinetic model to compute filtration rate parameter of each kidney ($F_T$). (a) Concentration curves obtained by DP and CS reconstructions are plotted for patient-2. With the proposed DP reconstruction, AIF peak is conserved while noise in the AIF tail is removed.  Correlation and Bland-Altman plots for $F_T$ parameters of each kidneys of  all patients are  compared to a CS reconstruction with optimal regularization parameter for functional evaluation (${\lambda:0.0125}$) with the results from DP in (b) and  CS with ${\lambda:0.125}$ in (c). Filtration rate parameters of each kidney obtained by DP and optimum CS show good agreement with a correlation coefficient of $R^2=0.94$ and all values are within the range of 1.96SD.  The DP parameters achieved higher correlation, and better agreement with the parameters from the optimal CS reconstruction.}
\end{figure}
Measured MRI signal in aorta and kidneys were averaged over manually segmented masks. Then, concentrations of the CA in aorta ($C_{AIF}$) and kidneys ($C_{kidney}$) were calculated by measuring the shortening in $T_1$ value prior to the injection of CA (i.e. baseline value) using GRE signal equation. $T_1$ values of aorta and kidneys prior to the contrast agent injection were assumed to be 1.6 and 1.2 s, respectively. To compute tracer kinetic (TK) parameters of kidneys, a separable 2-compartment (2-CP) model introduced by Sourbron et al. \cite{sourbron2008mri} was employed by solving the nonlinear least-squares optimization: $\min _{TK(\phi)}\|C_{kidney}(t)-C_{AIF}(t)*h(t)\|_2^2$ where $h(t)=F_P \exp(-t/T_P) + F_T/T_P \exp(-t/T_P) * \exp(-t/T_T)$. Here, $F_P$ is the plasma flow, $F_T$ is the tubular flow, $T_P$ and $T_T$ are the transit time of the tracer in plasma and tubular compartments. Note that glomerular filtration rate (GFR) of the kidney is related to the parameter $F_T$ by  GFR $= F_T x $ volume of kidney parenchyma. Bland-Altman analysis was exploited to compare estimated $F_T$ values from the proposed DP method and the CS method where the regularization parameter $\lambda$ was optimized for functional analysis (not for optimized image quality). 
For a single subject (patient-2), concentration curves of contrast agent in the aorta and kidneys as a function of time are given in Figure~\ref{fig:curveFit_fig}a. Correlation and Bland-Altman analysis plots for estimated $F_T$ values obtained from DP and optimum CS reconstructions (CS$_{\lambda:0.0125}$), for all subjects, are depicted in Figure~\ref{fig:curveFit_fig}b. The optimum regularization parameter for CS (CS$_{\lambda:0.0125}$) was selected as the same value in~\cite{kurugol2020prospective}, which showed high correlation ($R^2=0.98$) with the GFR values measured by nuclear medicine (NM-GFR).

The results show that the parameters estimated with the DP correlated well with the parameters from the reference CS reconstruction achieved with a lower $\lambda$ optimized for functional analysis (not for optimized image quality). When using a CS reconstruction with the largest $\lambda$, we could not achieve any reasonable $F_T$ values. When using the second largest $\lambda$, the estimated $F_T$ values did not correlate as well with the reference parameters compared to those from the DP reconstruction. 
\section{Discussion}
\label{sec:discuss}

There is a tradeoff when selecting the optimal hyperparameter ($\lambda$) for regularization in compressed sensing (CS) reconstruction. Selecting a larger hyperparameter value reduces the noise and the streaking artifacts due to undersampling, but it comes at the cost of removing the important temporal features (such as arterial input function peak) in the dynamic signal and degrading the accuracy of functional analysis. The proposed deep learning method combines the strength of multiple hyperparameters of CS as shown in Section~\ref{sec:Experiments}. Visually the proposed approach suppresses noise like a large hyperparameter CS reconstruction while at the same time not staggering from over-smoothing the temporal structure. Most importantly, the contrast effects are still apparent with DP approach without the expense of keeping artifacts. It is possible to explain this behavior by the merit of DP approach regularizing the temporal structure differently across patches, whereas the CS reconstruction smooths the temporal structure equally on the entire image. This effect is apparent especially for high $\lambda$ values posing a hyperparameter selection problem. 

The results in Figure~\ref{fig:curveFit_fig}a, showing concentration curves for DP and CS reconstructions, are consistent with the aforementioned conclusion since noise in the AIF tail, observed in CS$_{\lambda:0.0125}$ and lower $\lambda$ values, were removed but the AIF peak was preserved in DP. Although noise in the AIF tail was smoothed in CS$_{\lambda:0.125}$ and higher $\lambda$ values, the AIF peak was degraded. Correlation plots in Figure~\ref{fig:curveFit_fig}b, indicates a good agreement between optimum CS hyperparameter (CS$_{\lambda:0.0125}$) and DP with $R^2=0.94$. Furthermore, Bland-Altman plot in Figure~\ref{fig:curveFit_fig}b demonstrates that all values are within the range of 1.96 times of standard deviation. Comparing optimum CS (CS$_{\lambda:0.0125}$) and CS$_{\lambda:0.125}$ in Figure~\ref{fig:curveFit_fig}c, the correlation coefficient of estimated $F_T$ values is $R^2=0.90$ with all except two values within the range of 1.96SD. This is because CS reconstruction with a larger regularization parameter cannot preserve the peak of the AIF and results in inaccurate estimation of $F_T$ values.  

Reconstructing medical images with a regularization network has shown merit in several applications in PET imaging in \cite{gong2018pet}, CT imaging in \cite{gong2019low} an dynamic cardiac imaging in \cite{yoo2021time}. All previous contributions consider either single-image reconstruction or non-contrast dynamic imaging applications. In DCE-MRI dynamic structure of the image is determined by the contrast injection and ultimately this should be considered when designing a temporally correlated lower dimensional input sequence ($z_t$). 
One of our main contributions in this work is to design a suitable lower dimensional input sequence for DCE-MRI reconstruction. We experimented with different ways to design $z_t$. When we used a design in the form of a line, the resultant reconstructed images created an unrealistic representation of the temporal signal. When we selected a lower dimensional sequence in the form of a periodic signal, the resultant images were corrupted by noise enhancement and the important AIF peak was temporally smoothed, yielding unrealistic and inaccurate parameter estimations. Finally, we  introduced a simple piece-wise linear $z_t$ design where each linear phase is determined using principal components of temporal signal as explained in Section~\ref{subsec:lower_dim}. This approach yielded the reported results achieving both high quality images without undersampling artifacts and accurate temporal signal and improved image quality at the same time.

Another main advantage of the proposed method is its flexibility when compared to the standard supervised deep learning based image reconstruction algorithms, which often lack generalizability. Standard supervised algorithms are limited to datasets which are similar to the ones used in the training.  The proposed method on the other hand can be applied to any DCE-MRI data from different scanners, using different parameters, acquired with different coils, FOV or from different region or organs of interest, and from patients with different age groups. One limitation of this approach is that the run time is slower ($\sim$5 hours on GPU) compared to conventional reconstruction algorithms ($\sim$20-30min on CPU) due to single image-based training depending on the hardware configuration. 
\section{Conclusion}
\label{sec:conclusion}

In this manuscript we proposed a DIP approach for abdominal DCE-MR image reconstruction from undersampled golden angle radial acquisition. The goal is to both improve image quality, reducing streaking artifacts due to undersampling and also to keep the dynamic features of the temporal signal for quantitative analysis and estimation of tracer kinetic model parameters. Proposed approach removes the necessity of training with a large set of input and ground truth image pairs, as in classical supervised training. Through extensive evaluation of the proposed DIP based image reconstruction algorithm we observed improved performance in terms of both image quality and the ability to keep the temporal features of the dynamic sequence in aorta and kidney regions. The proposed method also improved the accuracy in estimating the filtration rate parameters of the kidneys  by fitting a two-compartment tracer kinetic model, when compared to CS reconstruction. {As a future work, we will investigate few shot transfer learning approaches to utilize an already learned regularization network for unseen data. Ultimately this will allow us to reduce time requirements and moreover can lead to extract correlations on different body part MRI scans.}

\tiny{
\bibliographystyle{IEEEtran}
\bibliography{ref}}

\begin{thebibliography}{10}
\providecommand{\url}[1]{#1}
\csname url@samestyle\endcsname
\providecommand{\newblock}{\relax}
\providecommand{\bibinfo}[2]{#2}
\providecommand{\BIBentrySTDinterwordspacing}{\spaceskip=0pt\relax}
\providecommand{\BIBentryALTinterwordstretchfactor}{4}
\providecommand{\BIBentryALTinterwordspacing}{\spaceskip=\fontdimen2\font plus
\BIBentryALTinterwordstretchfactor\fontdimen3\font minus
  \fontdimen4\font\relax}
\providecommand{\BIBforeignlanguage}[2]{{%
\expandafter\ifx\csname l@#1\endcsname\relax
\typeout{** WARNING: IEEEtran.bst: No hyphenation pattern has been}%
\typeout{** loaded for the language `#1'. Using the pattern for}%
\typeout{** the default language instead.}%
\else
\language=\csname l@#1\endcsname
\fi
#2}}
\providecommand{\BIBdecl}{\relax}
\BIBdecl

\bibitem{kurugol2020prospective}
S.~Kurugol, O.~Afacan, R.~S. Lee, C.~M. Seager, M.~A. Ferguson, D.~R. Stein,
  R.~C. Nichols, M.~Dugan, A.~Stemmer, S.~K. Warfield \emph{et~al.},
  ``Prospective pediatric study comparing glomerular filtration rate estimates
  based on motion-robust dynamic contrast-enhanced magnetic resonance imaging
  and serum creatinine (egfr) to 99m tc dtpa,'' \emph{Pediatric radiology},
  vol.~50, no.~5, pp. 698--705, 2020.

\bibitem{kurugol2020feed}
S.~Kurugol, C.~M. Seager, H.~Thaker, J.~Coll-Font, O.~Afacan, R.~C. Nichols,
  S.~K. Warfield, R.~S. Lee, and J.~S. Chow, ``Feed and wrap magnetic resonance
  urography provides anatomic and functional imaging in infants without
  anesthesia,'' \emph{Journal of pediatric urology}, vol.~16, no.~1, pp.
  116--120, 2020.

\bibitem{block2007undersampled}
K.~T. Block, M.~Uecker, and J.~Frahm, ``Undersampled radial mri with multiple
  coils. iterative image reconstruction using a total variation constraint,''
  \emph{Magnetic Resonance in Medicine: An Official Journal of the
  International Society for Magnetic Resonance in Medicine}, vol.~57, no.~6,
  pp. 1086--1098, 2007.

\bibitem{block2014towards}
K.~T. Block, H.~Chandarana, S.~Milla, M.~Bruno, T.~Mulholland, G.~Fatterpekar,
  M.~Hagiwara, R.~Grimm, C.~Geppert, B.~Kiefer \emph{et~al.}, ``Towards routine
  clinical use of radial stack-of-stars 3d gradient-echo sequences for reducing
  motion sensitivity,'' \emph{Journal of the Korean Society of Magnetic
  Resonance in Medicine}, vol.~18, no.~2, pp. 87--106, 2014.

\bibitem{ravishankar2010mr}
S.~Ravishankar and Y.~Bresler, ``Mr image reconstruction from highly
  undersampled k-space data by dictionary learning,'' \emph{IEEE transactions
  on medical imaging}, vol.~30, no.~5, pp. 1028--1041, 2010.

\bibitem{zhang2015accelerating}
T.~Zhang, J.~M. Pauly, and I.~R. Levesque, ``Accelerating parameter mapping
  with a locally low rank constraint,'' \emph{Magnetic resonance in medicine},
  vol.~73, no.~2, pp. 655--661, 2015.

\bibitem{hu2020multi}
Y.~Hu, X.~Wang, Q.~Tian, G.~Yang, B.~Daniel, J.~McNab, and B.~Hargreaves,
  ``Multi-shot diffusion-weighted mri reconstruction with magnitude-based
  spatial-angular locally low-rank regularization (spa-llr),'' \emph{Magnetic
  resonance in medicine}, vol.~83, no.~5, pp. 1596--1607, 2020.

\bibitem{otazo2015low}
R.~Otazo, E.~Candes, and D.~K. Sodickson, ``Low-rank plus sparse matrix
  decomposition for accelerated dynamic mri with separation of background and
  dynamic components,'' \emph{Magnetic resonance in medicine}, vol.~73, no.~3,
  pp. 1125--1136, 2015.

\bibitem{ong2016beyond}
F.~Ong and M.~Lustig, ``Beyond low rank+ sparse: Multiscale low rank matrix
  decomposition,'' \emph{IEEE journal of selected topics in signal processing},
  vol.~10, no.~4, pp. 672--687, 2016.

\bibitem{feng2016xd}
L.~Feng, L.~Axel, H.~Chandarana, K.~T. Block, D.~K. Sodickson, and R.~Otazo,
  ``Xd-grasp: golden-angle radial mri with reconstruction of extra motion-state
  dimensions using compressed sensing,'' \emph{Magnetic resonance in medicine},
  vol.~75, no.~2, pp. 775--788, 2016.

\bibitem{feng2014golden}
L.~Feng, R.~Grimm, K.~T. Block, H.~Chandarana, S.~Kim, J.~Xu, L.~Axel, D.~K.
  Sodickson, and R.~Otazo, ``Golden-angle radial sparse parallel mri:
  combination of compressed sensing, parallel imaging, and golden-angle radial
  sampling for fast and flexible dynamic volumetric mri,'' \emph{Magnetic
  resonance in medicine}, vol.~72, no.~3, pp. 707--717, 2014.

\bibitem{schlemper2017deep}
J.~Schlemper, J.~Caballero, J.~V. Hajnal, A.~N. Price, and D.~Rueckert, ``A
  deep cascade of convolutional neural networks for dynamic mr image
  reconstruction,'' \emph{IEEE transactions on Medical Imaging}, vol.~37,
  no.~2, pp. 491--503, 2017.

\bibitem{liu2021magnetic}
F.~Liu, R.~Kijowski, G.~El~Fakhri, and L.~Feng, ``Magnetic resonance parameter
  mapping using model-guided self-supervised deep learning,'' \emph{Magnetic
  Resonance in Medicine}, vol.~85, no.~6, pp. 3211--3226, 2021.

\bibitem{ke2020deep}
Z.~Ke, W.~Huang, J.~Cheng, Z.~Cui, S.~Jia, H.~Wang, X.~Liu, H.~Zheng, L.~Ying,
  Y.~Zhu \emph{et~al.}, ``Deep low-rank prior in dynamic mr imaging,''
  \emph{arXiv preprint arXiv:2006.12090}, 2020.

\bibitem{han2018deep}
Y.~Han, J.~Yoo, H.~H. Kim, H.~J. Shin, K.~Sung, and J.~C. Ye, ``Deep learning
  with domain adaptation for accelerated projection-reconstruction mr,''
  \emph{Magnetic resonance in medicine}, vol.~80, no.~3, pp. 1189--1205, 2018.

\bibitem{ulyanov2018deep}
D.~Ulyanov, A.~Vedaldi, and V.~Lempitsky, ``Deep image prior,'' in
  \emph{Proceedings of the IEEE conference on computer vision and pattern
  recognition}, 2018, pp. 9446--9454.

\bibitem{yoo2021time}
J.~Yoo, K.~H. Jin, H.~Gupta, J.~Yerly, M.~Stuber, and M.~Unser,
  ``Time-dependent deep image prior for dynamic mri,'' \emph{IEEE Transactions
  on Medical Imaging}, 2021.

\bibitem{doneva2008automatic}
M.~Doneva and P.~B{\"o}rnert, ``Automatic coil selection for channel reduction
  in sense-based parallel imaging,'' \emph{Magnetic Resonance Materials in
  Physics, Biology and Medicine}, vol.~21, no.~3, pp. 187--196, 2008.

\bibitem{zhang2013coil}
T.~Zhang, J.~M. Pauly, S.~S. Vasanawala, and M.~Lustig, ``Coil compression for
  accelerated imaging with cartesian sampling,'' \emph{Magnetic resonance in
  medicine}, vol.~69, no.~2, pp. 571--582, 2013.

\bibitem{coll2021modeling}
J.~Coll-Font, O.~Afacan, J.~S. Chow, R.~S. Lee, S.~K. Warfield, and S.~Kurugol,
  ``Modeling dynamic radial contrast enhanced mri with linear time invariant
  systems for motion correction in quantitative assessment of kidney
  function,'' \emph{Medical Image Analysis}, vol.~67, p. 101880, 2021.

\bibitem{coll2020bulk}
J.~Coll-Font, O.~Afacan, J.~S. Chow, R.~S. Lee, A.~Stemmer, S.~K. Warfield, and
  S.~Kurugol, ``Bulk motion-compensated dce-mri for functional imaging of
  kidneys in newborns,'' \emph{Journal of Magnetic Resonance Imaging}, vol.~52,
  no.~1, pp. 207--216, 2020.

\bibitem{muckley_tkbnufft}
M.~J. Muckley, R.~Stern, T.~Murrell, and F.~Knoll, ``{TorchKbNufft}: A
  high-level, hardware-agnostic non-uniform fast {Fourier} transform,'' in
  \emph{ISMRM Workshop on Data Sampling \& Image Reconstruction}, 2020.

\bibitem{sourbron2008mri}
S.~P. Sourbron, H.~J. Michaely, M.~F. Reiser, and S.~O. Schoenberg,
  ``Mri-measurement of perfusion and glomerular filtration in the human kidney
  with a separable compartment model,'' \emph{Investigative radiology},
  vol.~43, no.~1, pp. 40--48, 2008.

\bibitem{gong2018pet}
K.~Gong, C.~Catana, J.~Qi, and Q.~Li, ``Pet image reconstruction using deep
  image prior,'' \emph{IEEE transactions on medical imaging}, vol.~38, no.~7,
  pp. 1655--1665, 2018.

\bibitem{gong2019low}
K.~Gong, K.~Kim, D.~Wu, M.~K. Kalra, and Q.~Li, ``Low-dose dual energy ct image
  reconstruction using non-local deep image prior,'' in \emph{2019 IEEE Nuclear
  Science Symposium and Medical Imaging Conference (NSS/MIC)}.\hskip 1em plus
  0.5em minus 0.4em\relax IEEE, 2019, pp. 1--2.

\end{thebibliography}

\end{document}